\def\year{2021}\relax
\documentclass[letterpaper]{article} 
\usepackage{aaai21}  
\usepackage{times}  
\usepackage{helvet} 
\usepackage{courier}  
\usepackage[hyphens]{url}  
\usepackage{graphicx} 
\usepackage{natbib}  
\usepackage{caption} 
\usepackage{amssymb}
\usepackage{amsmath}
\usepackage{romannum}
\DeclareMathAlphabet\mathbfcal{OMS}{cmsy}{b}{n}

\usepackage{booktabs}
\usepackage{multirow}
\usepackage{comment}
\usepackage[dvipsnames]{xcolor}
\usepackage{adjustbox}
\usepackage{subcaption}
\usepackage{nicefrac}

\usepackage{tabularx}
\newcommand\setrow[1]{\gdef\rowmac{#1}#1\ignorespaces}
\newcommand\clearrow{\global\let\rowmac\relax}
\clearrow
\usepackage[switch]{lineno}
\usepackage{pdfpages}

\definecolor{darkblue}{rgb}{0.0, 0.0, 0.55}
\usepackage{hyperref}
\hypersetup{
     colorlinks   = true,
     citecolor    = darkblue,
     urlcolor     = darkblue
}

\title{Proxy Synthesis: Learning with Synthetic Classes for Deep Metric Learning}
\author {
        Geonmo Gu\thanks{Authors contributed equally.}\textsuperscript{\rm 1},
        Byungsoo Ko\footnotemark[1]\textsuperscript{\rm 1},
        Han-Gyu Kim \textsuperscript{\rm 2} \\
}
\affiliations {
    \textsuperscript{\rm 1} NAVER/LINE Vision, 
    \textsuperscript{\rm 2} NAVER Clova Speech \\
    korgm403@gmail.com, kobiso62@gmail.com, hangyu.kim@navercorp.com \\
    \href{https://github.com/navervision/proxy-synthesis}{github.com/navervision/proxy-synthesis}
}

\begin{document}
\maketitle

\begin{abstract}
One of the main purposes of deep metric learning is to construct an embedding space that has well-generalized embeddings on both \textit{seen} (training) classes and \textit{unseen} (test) classes.
Most existing works have tried to achieve this using different types of metric objectives and hard sample mining strategies with given training data.
However, learning with only the training data can be overfitted to the \textit{seen} classes, leading to the lack of generalization capability on \textit{unseen} classes.
To address this problem, we propose a simple regularizer called \textit{Proxy Synthesis} that exploits synthetic classes for stronger generalization in deep metric learning.
The proposed method generates synthetic embeddings and proxies that work as synthetic classes, and they mimic \textit{unseen} classes when computing proxy-based losses.
\textit{Proxy Synthesis} derives an embedding space considering class relations and smooth decision boundaries for robustness on \textit{unseen} classes.
Our method is applicable to any proxy-based losses, including softmax and its variants.
Extensive experiments on four famous benchmarks in image retrieval tasks demonstrate that \textit{Proxy Synthesis} significantly boosts the performance of proxy-based losses and achieves state-of-the-art performance.
\end{abstract}

\section{Introduction}

Deep metric learning aims to learn a similarity metric among arbitrary data points so that it defines an embedding space where semantically similar images are close together, and dissimilar images are far apart.
Owing to its practical significance, it has been used for a variety of tasks such as image retrieval~\cite{gordo2016deep, sohn2016improved}, person re-identification~\cite{yu2018hard, hermans2017defense}, zero-shot learning~\cite{zhang2016zero}, and face recognition~\cite{wen2016discriminative, deng2019arcface}.
The well-structured embedding is requested to distinguish the \textit{unseen} classes properly, where the model is required to learn image representation from \textit{seen} classes.
This has been achieved by loss functions, which can be categorized into two types: \textit{pair-based} and \textit{proxy-based} loss.

The pair-based losses are designed based on the pair-wise similarity between data points in the embedding space, such as contrastive~\cite{chopra2005learning}, triplet~\cite{weinberger2009distance}, N-pair loss~\cite{sohn2016improved}, \textit{etc}.
However, they require high training complexity and empirically suffer from sampling issues~\cite{movshovitz2017no}.
To address these issues, the concept of proxy has been introduced.
A proxy is a representative of each class, which can be trained as a part of the network parameters.
Given a selected data point as an anchor, proxy-based losses consider its relations with proxies.
This alleviates the training complexity and sampling issues because only data-to-proxy relations are considered with a relatively small number of proxies compared to that of data points.

Although the performance of metric learning losses has been improved, a network trained only with training (\textit{seen}) data can be overfitted to the \textit{seen} classes and suffer from low generalization on \textit{unseen} classes.
To resolve this problem, previous works~\cite{zheng2019hardness, gu2020symmetrical, ko2020embedding} have generated synthetic samples to exploit additional training signals and more informative representations.
However, these methods can only be used for pair-based losses; thus, they still suffer from the training complexity and sampling issues.

\begin{figure*}[t!h!]
\centering
\includegraphics[width=0.85\textwidth]{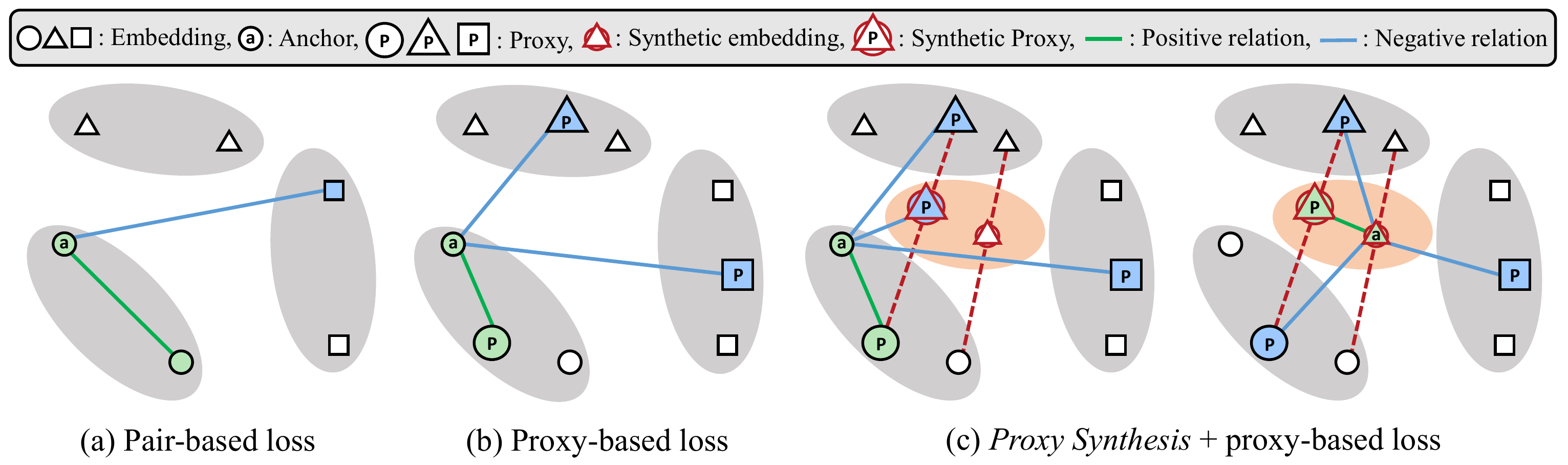}
\caption{Comparison among concepts of pair-based loss, proxy-based loss, and \textit{Proxy Synthesis} + proxy-based loss. (a) Pair-based loss maximizes similarity of positive pairs and minimizes similarity of negative pairs (i.e., Triplet loss). (b) Given an anchor embedding, proxy-based loss maximizes similarity with positive proxy and minimizes similarity with all negative proxies (i.e., Proxy NCA and Softmax variants). (c) \textit{Proxy Synthesis} exploits synthetic classes in-between original classes for additional training signals and competitive hard classes.
}
\label{fig:comparison}
\end{figure*}

In this paper, we propose \textit{\textbf{Proxy Synthesis}} (\textit{PS}), which is a simple regularizer for proxy-based losses that encourages networks to construct better generalized embedding space for \textit{unseen} classes.
As illustrated in Figure~\ref{fig:comparison}, our method generates synthetic embeddings and proxies as synthetic classes for computing a proxy-based loss.
\textit{Proxy Synthesis} exploits synthetic classes generated by semantic interpolations to mimic \textit{unseen} classes, obtaining smooth decision boundaries and an embedding space considering class relations.
Moreover, the proposed method can be used with any proxy-based loss, including softmax loss and its variants.
We demonstrate that our proposed method yields better robustness on \textit{unseen} classes and deformation on the input and embedding feature.
We achieve a significant performance boost on every proxy-based loss with \textit{Proxy Synthesis} and obtain state-of-the-art performance with respect to four famous benchmarks in image retrieval tasks.

\section{Related Work}
\label{sec:related_work}

\paragraph{Sample Generation:}
To achieve better generalization, previous works~\cite{zhao2018adversarial, duan2018deep, zheng2019hardness} have leveraged a generative network to create synthetic samples, which can lead to a bigger model and slower training speed.
To solve these problems, recent works~\cite{gu2020symmetrical, ko2020embedding} have proposed to generate samples by algebraic computation in the embedding space.
However, the above works can only be used for pair-based losses, which causes the same drawbacks of high training complexity and careful pair mining.
In addition, the above works exploit synthetic embeddings only for existing (\textit{seen}) classes, when \textit{Proxy Synthesis} uses synthetic embeddings and proxies as virtual classes for generalization on \textit{unseen} classes explicitly.

\paragraph{Mixup:}
Mixup techniques~\cite{zhang2017mixup,verma2018manifold,guo2019mixup} have been proposed for generalization in the classification task.
These techniques linearly interpolate a random pair of training samples and the corresponding one-hot labels.
\textit{Proxy Synthesis} and Mixup techniques share the common concept in terms of interpolating features for augmentation but have three major differences.
First, Mixup techniques are proposed for generalization, which aims for robustness on \textit{seen} classes, such as classification, whereas \textit{Proxy Synthesis} is proposed for generalization in metric learning tasks, aiming for robustness on \textit{unseen} classes.
Second, Mixup techniques interpolate the input vectors and hidden representations, whereas the proposed method interpolates the embedding features in the output space.
Third, Mixup techniques interpolate one-hot labels, while \textit{Proxy Synthesis} interpolates proxies, which allow us to learn the positional relations of class representatives in the embedding space explicitly.

\paragraph{Virtual Class:}
Virtual softmax~\cite{chen2018virtual} generates a single weight as a virtual negative class for softmax function to enhance the discriminative property of learned features in the classification task.
Even though the work proves that the constrained region for each class becomes more compact by the number of classes increase, Virtual softmax considers a single synthetic weight without any corresponding embedding as a virtual negative class.
Moreover, generating virtual weight by $W_{virt}=\nicefrac{\parallel W_{y_i}\parallel x_i}{\parallel x_i \parallel}$ is not applicable for softmax variants with weight normalization (i.e. Norm-Softmax, ArcFace, Proxy-anchor, \textit{etc}), where $x_i$ is $i$-th embedding, and $W_{y_i}$ is its positive class weight.
This is because $W_{virt}$ of the synthetic negative class will be equivalent to $x_i$ after normalization.
In contrast, \textit{Proxy Synthesis} generates multiple proxies (weights) and corresponding embeddings as multiple virtual classes, which can be used as negative and also positive classes.
Moreover, the proposed method is applicable for any proxy-based loss and softmax variants.

\section{Proposed Method}
\label{sec:proxy_synthesis}

\subsection{Preliminary}

Consider a deep neural network $f:\mathbfcal{D} \xrightarrow{f} \mathbfcal{X}$, which maps from an input data space $\mathbfcal{D}$ to an embedding space $\mathbfcal{X}$.
We define a set of embedding feature $X = [x_1, x_2, \dots, x_N]$, where each feature $x_i$ has label of $y_i \in \{1, \dots ,C\}$ and $N$ is the number of embedding features.
We denote a set of proxy $P = [p_1, p_2, \dots, p_C]$ and formulate generalized proxy-based loss as:
\begin{eqnarray}
\mathcal L(X,P) = \mathop{\mathbb{E}}_{(x,p) \sim R} \ell(x,p),
\end{eqnarray}
where $(x,p)$ denotes random pair of embedding and matching proxy from the pair distribution $R$.

Softmax loss is not only the most widely used classification loss but also has been re-valued as competitive loss in metric learning~\cite{zhai2018classification, boudiaf2020metric}.
Let $W_j \in \mathbb{R}^d$ denote the $j$-th column of the weight $W \in \mathbb{R}^{d \times C}$, where $d$ is the size of embedding features.
Then, Softmax loss is presented as follows:
\begin{eqnarray}
\mathcal{L}_{Softmax}(X) = -\frac{1}{\mid X \mid}\sum_{i=1}^{\mid X \mid} \log\frac{e^{W^T_{y_i}x_i}}{\sum_{j=1}^{C}e^{W^T_j x_i}},
\end{eqnarray}
where we set the bias $b=0$ because it does not affect the performance~\cite{liu2017sphereface, deng2019arcface}.
Because the proxy $P$ is learned as model parameter, the weight $W$ of softmax loss can be interpreted as proxy, which is the center of each class~\cite{deng2019arcface, wang2018cosface}.

Normalizing the weights and feature vector is proposed to lay them on a hypersphere of a fixed radius for better interpretation and performance~\cite{wang2017normface, wang2018additive, liu2017sphereface}.
When we transform the logit~\cite{pereyra2017regularizing} as $W_j^T x_i = {\parallel W_j \parallel \parallel x_i \parallel \cos{\theta_j}}$ and fix the individual proxy (weight) $\parallel W_j \parallel = 1$ and feature $\parallel x_i \parallel = 1$ by $l_2$-normalization, the normalized softmax (Norm-softmax) loss can be written with proxy-wise form as:
\begin{eqnarray}
\mathcal{L}_{Norm}(X,P) = -\frac{1}{\mid X \mid}\sum_{x \in X}\log\frac{e^{\gamma s(x,p^+)}}{e^{\gamma s(x,p^+)} + \sum\limits_{q \in P^-}e^{\gamma s(x,q)}},
\end{eqnarray}
where $p^+$ is a positive proxy, $P^-$ is a set of negative proxies, $\gamma$ is a scale factor, and $s(a,b)$ denotes the cosine similarity between $a$ and $b$.
More details of proxy-based (Proxy-NCA~\cite{movshovitz2017no}, SoftTriple~\cite{qian2019softtriple}, and Proxy-anchor~\cite{kim2020proxy}) and softmax variants (SphereFace~\cite{liu2017sphereface}, ArcFace~\cite{deng2019arcface}, and CosFace~\cite{wang2018cosface}) losses are presented in the supplementary Section A.

\subsection{Proxy Synthesis}

One of the key purposes of metric learning is to construct a robust embedding space for \textit{unseen} classes.
For this purpose, the proposed method allows proxy-based losses to exploit synthetic classes.
Training a proxy-based loss using \textit{Proxy Synthesis} is performed in three steps.
First, we process a mini-batch of input data with a network $f$ to obtain a set of embeddings $X$.
Second, given two random pairs of an embedding and corresponding proxy from different classes, $(x, p)$ and $(x', p')$, we generate a pair of synthetic embedding and proxy $(\Tilde{x}, \Tilde{p})$ as follows:
\begin{eqnarray}
(\Tilde{x}, \Tilde{p}) \equiv (I_{\lambda}(x,x'), I_{\lambda}(p,p')), \label{eq:gen}
\end{eqnarray}
where $I_{\lambda}(a,b)=\lambda a + (1-\lambda)b$ is a linear interpolation function with the coefficient of $\lambda \sim Beta(\alpha,\alpha)$ for $\alpha \in (0,\infty)$, and $\lambda \in [0,1]$.
We perform $\mu \times batch\,size$ generations of Eq.~\ref{eq:gen}, where hyper-parameter $\mu = \frac{\#\,of\,synthetics}{batch\,size}$ is a generation ratio by batch size.
Thereafter, we define $\widehat{X}$ as a set of original and synthetic embeddings, and $\widehat{P}$ as a set of original and synthetic proxies.
Synthetic proxy $\Tilde{p}$ will work as a representative of the synthetic class, which has a mixed representation of class $p$ and $p'$, whereas the synthetic embedding $\Tilde{x}$ will be a synthetic data point of the synthetic class.
Third, we compute the loss, including synthetic embeddings and proxies, as if they are new classes.
The generalized loss with \textit{Proxy Synthesis} is formulated as:
\begin{eqnarray}
\mathcal L(\widehat{X},\widehat{P}) = \mathop{\mathbb{E}}_{\lambda \sim Beta(\alpha, \alpha)}\mathop{\mathbb{E}}_{(x,p) \sim \widehat{R_\lambda}}\ell(x,p), \label{eq:ps}
\end{eqnarray}
where $\widehat{R_\lambda}$ is a distribution of the embedding and proxy pairs including originals and synthetics generated with $\lambda$.
Implementing \textit{Proxy Synthesis} is extremely simple with few lines of codes.
Moreover, it does not require to modify any code of proxy-based loss and can be used in a plug-and-play manner with negligible computation cost.
Code-level description and experiment of training time and memory are presented in the supplementary Section B.1 and D.1, respectively.

\begin{figure*}[h!t!]
\centering
\includegraphics[width=0.75\textwidth]{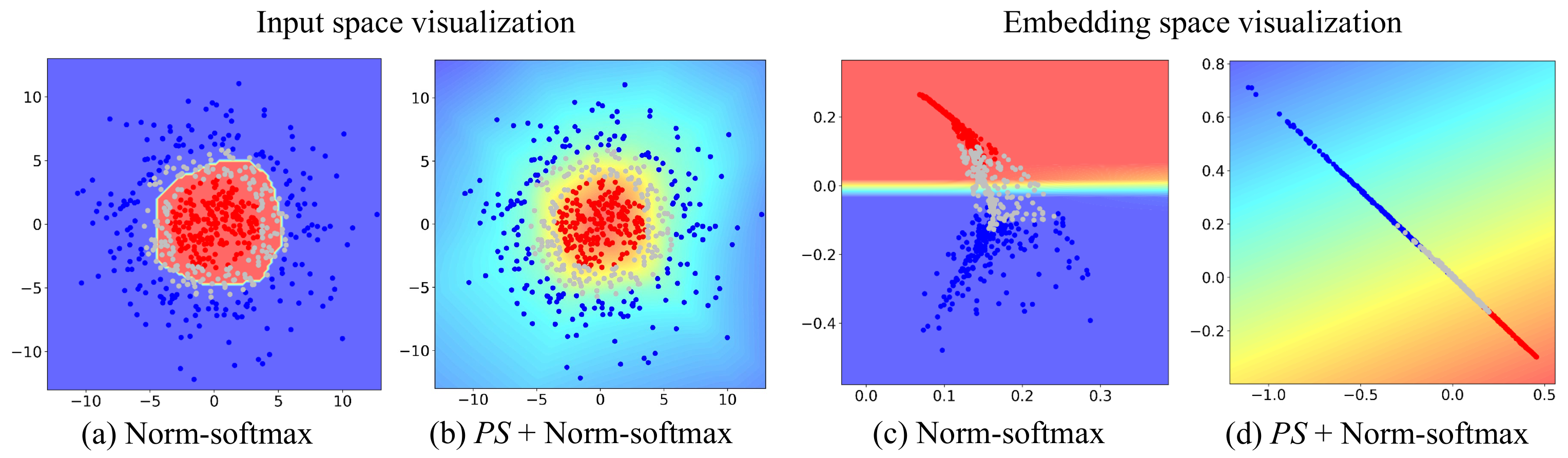}
\caption{Experiment with 2D isotropic Gaussian dataset including red, blue, and gray classes. Simple feed-forward network with two-dimensional embedding is used, while we train network with red and blue classes and let gray class remain \textit{unseen}. The darker the intensity of the blue or red in the background, the higher the prediction confidence to blue or red class, respectively.}
\label{fig:space_vis}
\end{figure*}
 
\subsection{Discussion}

\subsubsection{Learning with Class Relations:}

Unlike tasks that only test with \textit{seen} classes such as classification, metric learning is desired to construct a structural embedding space for robustness on \textit{unseen} classes.
A well-structured embedding space should contain meaningful relations among embeddings; an example from the previous work~\cite{mikolov2013efficient} is as follows: $king - man + woman \approx queen$ in a word embedding space.
To achieve this, \textit{Proxy Synthesis} explicitly inserts in-between class relations with synthetic classes (i.e., $I_\lambda(wolf, dog) \approx wolfdog$) and they mimic unseen classes for training with a diverse range of data characteristics.
\textit{Proxy Synthesis} considers class relations with Equation~\ref{eq:gen} and~\ref{eq:ps} in forward propagation.
This characteristic is reflected in backward propagation as well.

For convenience in gradient analysis, we write the loss of softmax function on $(x,p_i)$, where $x$ is an anchor embedding of input, and $p_i$ is corresponding positive proxy, as follows,
\begin{eqnarray}
\mathcal{L}_i  =  \mathcal{L}_{Softmax}(x,p_i)  =   -\log \frac{E(p_i)}{E(p_i) + \sum_{q \in P^-}E(q)},
\label{eq:softmaxloss}
\end{eqnarray}
where $E(p) = e^{S(x,p)}$ and $S(x,p) = s(x,p) \parallel x \parallel \parallel p \parallel = x^T p$. Then, gradient over positive similarity $S(x,p_i)$ is,
\begin{eqnarray}
\frac{\partial \mathcal{L}_i}{\partial S(x,p_i)} & = &  \frac{E(p_i)}{\sum_{q \in P}E(q)} - 1.
\label{eq:softmaxgradient}
\end{eqnarray}
It shows that the gradient over $S(x,p_i)$ only considers the similarity of the anchor embedding and its proxy by $E(p_i)$.

When \textit{Proxy Synthesis} is applied, the gradient changes. In this gradient induction, we assume that the positive proxy $p_i$ of input is used for generating synthesized proxy $\tilde{p}$ with $p_j$ as $\Tilde{p}=\lambda p_i + \lambda' p_j$, where $\lambda'=1-\lambda$. Then, the gradients over $S(x,p_i)$ and $S(x,p_j)$ are inducted as follows,
\begin{eqnarray}
\frac{\partial \mathcal{L}_i}{\partial S(x,p_i)} & = & \frac{\lambda E(\tilde{p}) + E(p_i)}{E(\tilde{p}) + \sum_{q \in P}E(q)} - 1, \\ 
\frac{\partial \mathcal{L}_i}{\partial S(x,p_j)} & = & \frac{\lambda' E(\tilde{p}) + E(p_j)}{E(\tilde{p}) + \sum_{q \in P}E(q)}. \label{eq:ps_g}
\end{eqnarray}
In contrast to the softmax loss, \textit{Proxy Synthesis} enables the gradient over $S(x,p_i)$ and $S(x,p_j)$ to consider class relation between $p_i$ and $p_j$ via $E(\tilde{p}) = E(\lambda p_i + \lambda' p_j)$ in the backward propagation.
The detailed induction is presented in supplementary Section B.2.

\subsubsection{Obtaining a Smooth Decision Boundary:}
Synthetic classes work as hard competitors of original classes because of positional proximity, which leads to lower prediction confidence and, thus, smoother decision boundaries.
The smoothness of the decision boundary is a main factor of generalization~\cite{bartlett1999generalization, verma2018manifold}, and it is more desirable in metric learning to provide a relaxed estimate of uncertainty for \textit{unseen} classes.
For better intuitions, we conduct an experiment to visualize the generalization effect of \textit{Proxy Synthesis}, as depicted in Figure~\ref{fig:space_vis}.
For both the input and embedding spaces, Norm-softmax has a strict decision boundary, whereas \textit{PS} + Norm-softmax has a smooth decision boundary that transitions linearly from the red to the blue class.
Further theoretical analysis of such phenomenon is provided below.

A network with ordinary softmax outputs the probability of an embedding $x$ belonging to a specific class $i$ as follows,
\begin{eqnarray}
\texttt{Pr}(x, i) = \frac{e^{S(x, p_i)}}{e^{S(x, p_i)} + \sum_{q \in P \setminus \{p_i\}} e^{S(x, q)}}.
\end{eqnarray}
Considering the linearity of similarity function $S(x,p) = x^T p$, the strict decision boundary is constructed when the network always outputs embedding vector close to one proxy, even though the input has shared visual semantics among different classes.
Such phenomenon occurs because the softmax function forces all embedding vectors to become as close to the corresponding proxies as possible during training.
Considering the softmax loss of an embedding vector and its positive proxy $(x, p_i)$ as Equation~\ref{eq:softmaxloss}, the gradient of such loss over $x$ can be inducted as follows:
\begin{eqnarray}
\frac{\partial\mathcal{L}_i}{\partial x} = \tau_i p_i & + & \sum_{p_k \in P^-}{\tau_k p_k}, \\
\tau_i = \frac{E(p_i)}{\sum_{q \in P}E(q)} - 1 & , & \tau_k =   \frac{E(p_k)}{\sum_{q \in P}E(q)}.
\end{eqnarray}
It is obvious that $\tau_i < 0$ and $\tau_k > 0$.
Considering the parameter update is performed by $x = x - \eta \frac{\partial\mathcal{L}_i}{\partial x}$, where $\eta$ is a learning rate, the gradient descent forces $x$ to be closer to $p_i$ and to be distant from other proxies $p_k$.

\textit{Proxy Synthesis} overcomes such problem of embedding space to overfit to the proxies by providing a gradient of opposite direction compared to ordinary softmax.
To describe major difference of \textit{Proxy Synthesis} and ordinary softmax, we consider softmax loss for synthesized pair  $\Tilde{x}=\lambda x + \lambda' x', \Tilde{p}=\lambda p_i + \lambda' p_j$, where $(x, p_i)$ and $(x', p_j)$ are pairs of an embedding and a corresponding proxy:
\begin{eqnarray}
\tilde{\mathcal{L}} & = & \mathcal{L}_{Softmax}(\tilde{x},\tilde{p}) \\
                & = &  -\log \frac{E(\tilde{p})}{E(\tilde{p}) + E(p_i) + E(p_j) + \sum_{q \in P^-}E(q)} \nonumber,
\end{eqnarray}
where $E(p) = e^{S(\tilde{x},p)}$ and $P^- = P \setminus \{ p_i, p_j\}$. It should be noted that $\tilde{p} \notin P$.
Since we suggest to sample $\lambda$ from $Beta(\alpha,\alpha)$ with small $\alpha$ in Section~\ref{sec:hardness}, $\tilde{p}$ has high chance to be generated either close to $p_i$ with $\lambda >> 0.5$ or close to $p_j$ with $\lambda << 0.5$.
As the proofs for both cases are equivalent, we assume the first case: $\tilde{x}$ is much closer to $x$ than $x'$.
We consider gradient over $x$ because the loss will affect the closer embedding vector more than the other one. 
The inducted gradient over $x$ is as follows:
\begin{eqnarray}
\frac{\partial\tilde{\mathcal{L}}}{\partial x} & = & - \lambda \frac{\sum_{q \in P}(\tilde{p} - q)E(q)}{E(\tilde{p}) + \sum_{q \in P}E(q)}.
\label{eq:synpartial_1}
\end{eqnarray}
As $\tilde{x}$ is closer to $p_i$ and $\tilde{p}$ compared to other proxies, $E(p_i), E(\tilde{p}) >> E(q) \quad \forall q \neq p_i, \tilde{p}$.
Thus, Equation~\ref{eq:synpartial_1} can be re-written as follows,
\begin{eqnarray}
\frac{\partial\tilde{\mathcal{L}}}{\partial x}  \approx  - \lambda \frac{ (\tilde{p} - p_i) E(p_i) }{E(\tilde{p})+E(p_i)} =  \tau'_i p_i + \tau'_j p_j,\\
\tau'_i  =  \frac{\lambda \lambda'E(p_i)}{E(\tilde{p})+E(p_i)}, \quad \tau'_j  =  - \frac{\lambda \lambda'E(p_i)}{E(\tilde{p})+E(p_i)}.
\end{eqnarray}
It is obvious that $\tau'_i>0$, implying that by adopting \textit{Proxy Synthesis}, softmax loss for synthesized pair provides gradient which leads embedding vector not too close to the corresponding proxy $p_i$; it is also obvious that $\tau'_j<0$, implying that softmax loss for synthesized pair provides gradient which makes embedding vector not too distant from the competing proxy $p_j$.
In such a manner, \textit{Proxy Synthesis} prevents embedding vectors lying too close to proxies, which finally leads to the smooth decision boundary.
The detailed induction is provided in the supplementary Section B.3.

\begin{figure*}[!htp]
\centering
\includegraphics[width=0.75\textwidth]{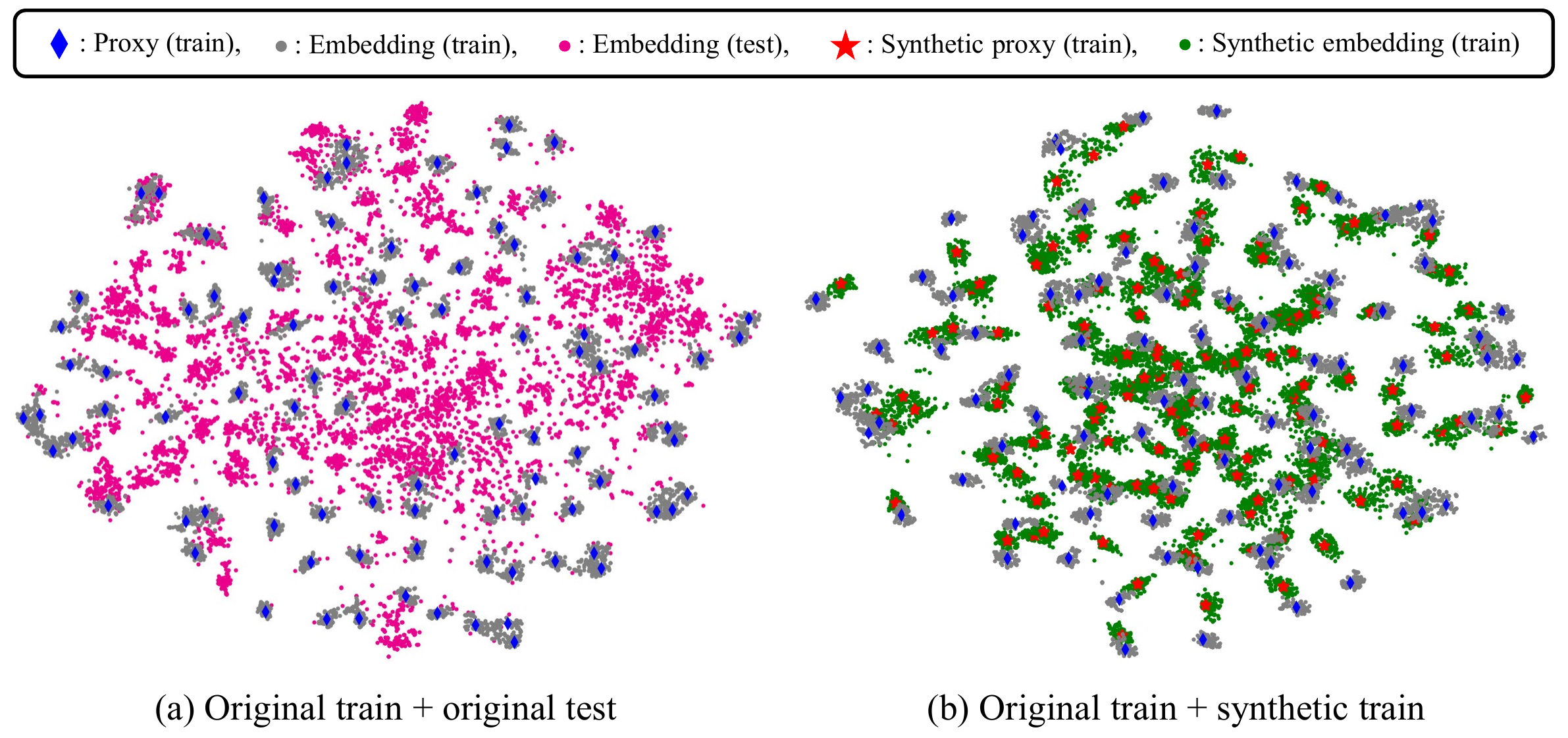}
\caption{t-SNE visualization~\cite{maaten2008visualizing} of converged network trained with \textit{PS} + Norm-softmax loss on CARS196. (a) We project both train (\textit{seen}) and test (\textit{unseen}) embeddings. (b) With the same train embeddings as in (a), we project synthetic embeddings and proxies.}
\label{fig:tsne}
\end{figure*}

\section{Experiments}
\label{sec:experiments}

\subsection{Experimental Setting}
\label{sec:experimental_setting}

We evaluate the proposed method with respect to four benchmarks in metric learning: CUB-200-2011 (CUB200)~\cite{wah2011caltech}, CARS196~\cite{krause20133d}, Standford Online Products (SOP)~\cite{oh2016deep}, and In-Shop Clothes (In-Shop)~\cite{liu2016deepfashion}.
We follow the widely used training and evaluation procedure from~\cite{oh2016deep,kim2020proxy} and call it \textit{conventional evaluation}.
Experiments are performed on an Inception network with batch normalization~\cite{ioffe2015batch} with a 512 embedding dimension.
For the hyper-parameters of \textit{Proxy Synthesis}, $\alpha$ and $\mu$ are set to 0.4 and 1.0, respectively.
Considering recent works~\cite{musgrave2020metric,fehervari2019unbiased} that have presented enhanced evaluation procedure with regard to fairness, we include an evaluation procedure designed from work ``A metric learning reality check''~\cite{musgrave2020metric} and call it \textit{MLRC evaluation}, which contains 4-fold cross-validation, ensemble evaluation, and usage of fair metrics (P@1, RP, and MAP@R).
Please refer to supplementary Section C for further details on the benchmarks and implementation.

\begin{table}[t]
\begin{adjustbox}{width=1.0\columnwidth,center}
\begin{tabular}{lccccc}
\toprule
\multirow{2}{*}{Model} & \multicolumn{2}{c}{Embedding} & \multicolumn{2}{c}{Proxy} & \multirow{2}{*}{R@1} \\ \cmidrule(r){2-5}
                      & Original      & Synthetic      & Original    & Synthetic    &                           \\ \midrule
M1 (baseline)          & \checkmark             &               & \checkmark           &             & 83.3                     \\
M2                     &              & \checkmark              &            & \checkmark            & 83.1                     \\
M3                     & \checkmark             &               & \checkmark           & \checkmark            & 83.7                     \\
M4                     &              & \checkmark              & \checkmark           & \checkmark            & 83.7                     \\
\textit{Proxy Synthesis} & \checkmark             & \checkmark              & \checkmark           & \checkmark            & \textbf{84.7}                     \\\bottomrule
\end{tabular}
\end{adjustbox}
\caption{Recall@1(\%) comparison among different usages of original and synthetic embedding and proxy on CARS196. We set elements of $\widehat{X}$ and $\widehat{P}$ to be checked(\checkmark) embeddings and proxies to compute $\mathcal{L}_{Norm}(\widehat{X},\widehat{P})$.}
\label{table:synthetics}
\end{table}

\subsection{Impact of Synthetic Class}
\label{sec:impact}

\subsubsection{Embedding Space Visualization:}
Exploiting synthetic classes is preferable in metric learning because the main goal is to develop robustness on \textit{unseen} classes.
This is depicted visually in Figure~\ref{fig:tsne}.
In Figure~\ref{fig:tsne}a, \textit{unseen} test embeddings are located in-between the clusters of train embeddings by forming clusters.
Similarly, synthetic classes are also generated in-between train embeddings, as depicted in Figure~\ref{fig:tsne}b, and play an important role in mimicking \textit{unseen} classes during the training phase.
Thus, these additional training signals enable a network to capture extra discriminative features for better robustness on \textit{unseen} classes.
Extended visualization is in supplementary Section D.5.

\subsubsection{Impact of Synthetic Embedding and Proxy:}
To investigate the quantitative impact of synthetic embedding and proxy, we conduct an experiment by differentiating the elements of $\widehat{X}$ and $\widehat{P}$ in Norm-softmax loss.
Table~\ref{table:synthetics} illustrates that using only synthetic embeddings and proxies (M2) leads to a slightly lower performance than the baseline (M1).
Adding synthetic proxies (M3) and using synthetic embedding instead of the original embedding (M4) leads to improved performance when compared with the baseline (M1).
This indicates that the generated synthetic embeddings and proxies build meaningful virtual classes for training.
Finally, using all embeddings and proxies (\textit{Proxy Synthesis}) achieves the best performance among all cases by considering the fundamental and additional training signals.

\begin{table}[t]
\begin{subtable}[h]{0.48\columnwidth}
\begin{adjustbox}{width=0.48\columnwidth,center}
\centering
\begin{tabular}{lc}
\toprule
$\lambda$ & R@1(\%)  \\
\midrule
0.1    & 83.1 \\
0.2    & \textbf{83.8} \\
0.3    & 83.7 \\
0.4    & 83.5 \\
0.5    & 83.3 \\
\bottomrule
\end{tabular}
\end{adjustbox}
\caption{Static generation}
\label{table:interpolation1}
\end{subtable}
\hfill
\begin{subtable}[h]{0.48\columnwidth}
\begin{adjustbox}{width=0.48\columnwidth,center}
\centering
\begin{tabular}{lc}
\toprule
$\alpha$ & R@1(\%)           \\
\midrule
0.2   & 84.0          \\
0.4   & \textbf{84.7} \\
0.8   & 83.9          \\
1.0   & 83.7          \\
1.5   & 83.7         \\
\bottomrule
\end{tabular}
\end{adjustbox}
\caption{Stochastic generation}
\label{table:interpolation2}
\end{subtable}
\caption{Comparison between static and stochastic generation of synthetics while training with \textit{PS} + Norm-softmax on CARS196. For static generation, synthetics are generated with fixed value of $\lambda$. For stochastic generation, synthetics are generated with sampled $\lambda$ from $Beta(\alpha, \alpha)$.}
\label{table:interpolation}
\end{table}

\begin{figure}[t]
\centering
\includegraphics[width=1.0\columnwidth]{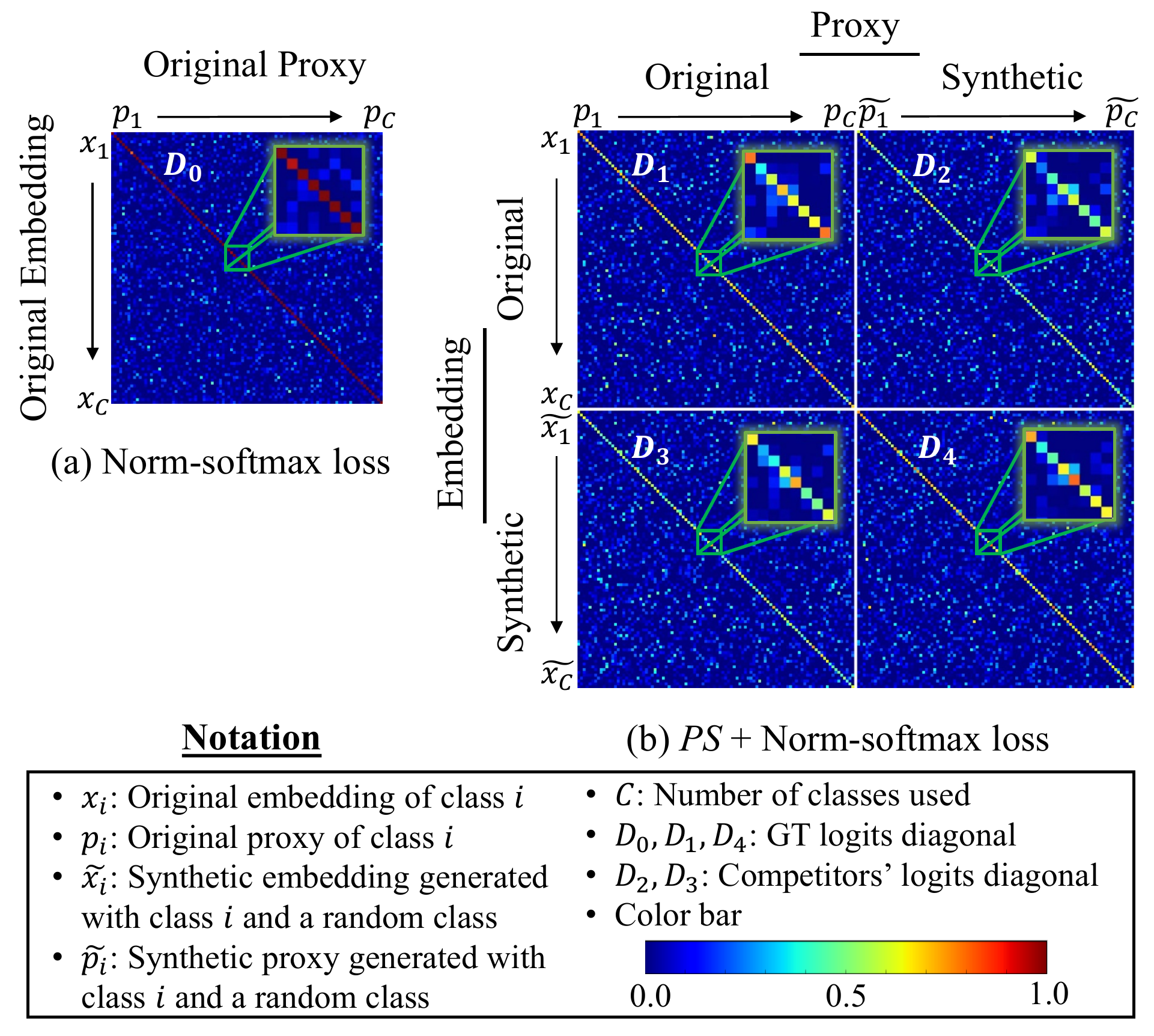}
\caption{Heatmap visualization of cosine similarity (logit) between embeddings and proxies at 100th epoch of training on CARS196. (a) Norm-softmax loss with original embedding and proxy. (b) \textit{PS} + Norm-softmax loss including synthetic embedding and proxy.}
\label{fig:logits_vis}
\end{figure}

\subsection{Synthetic Class as Hard Competitor}
\label{sec:hardness}

\subsubsection{Impact of Hardness:}
Generating synthetic classes is required to be hard enough so that the model can learn more discriminative representations.
The hardness of synthetic classes can be controlled by $\alpha$, which decides probability distribution for the sampling of the interpolation coefficient $\lambda$.
In Table~\ref{table:interpolation}, static $\lambda=0.1$ shows low performance because synthetic classes are too close to original classes, and static $\lambda=0.5$ also shows low performance because it generates synthetic classes in the middle of two original classes, which is relatively easy to distinguish.
The optimal static $\lambda$ value is around 0.2, which establishes the proper difficulty of distinguishment.
Moreover, the result shows that the stochastic generation is better than the static generation.
This is because stochastic generation can generate more number of different synthetic classes with wide variation.
In the stochastic generation, $\alpha=1.0$ is the same with uniform distribution, and $\alpha=1.5$ has a high chance of generating synthetics in the middle of two classes; thus, their performance is relatively low.
Similar to the experiment of static generation, $\alpha$ around $0.4$ shows the best performance, which has a high chance of generating synthetic classes close enough to an original class.
We provide additional experiments on the effect of hyper-parameter in supplementary Section D.2.

\subsubsection{Logits Visualization:}
We compare the cosine similarity (logits) between embeddings and proxies during the training procedure.
In Figure~\ref{fig:logits_vis}a, the logit values of ground truth (GT), which are represented by the main diagonal $D_0$, are clearly red owing to high prediction confidence.
This leads to a strict decision boundary, as depicted in Figure~\ref{fig:space_vis}a and Figure~\ref{fig:space_vis}c, and may cause an overfitting problem.
On the other hand, the GT logit values of \textit{PS} + Norm-softmax ($D_1$ and $D_4$) have lower confidence, represented with yellow to orange color.
This is because synthetic classes generated near the original classes work as hard competitors ($D_2$ and $D_3$), which prevents excessively high confidence, while the confidence of main diagonals ($D_1$ and $D_4$) is still higher than that of competitors' diagonals ($D_2$ and $D_3$) with redder color for the same embedding.
This smoothens the decision boundary, as depicted in Figure~\ref{fig:space_vis}b and Figure~\ref{fig:space_vis}d, and leads to stronger generalization.

\begin{table}[t]
\begin{adjustbox}{width=0.75\columnwidth,center}
\begin{tabular}{lcc}
\toprule
Deformation           & Norm-softmax   & \textit{PS} + Norm-softmax \\
\midrule
Cutout   & 75.3 & 77.0 \textcolor{Green}{(+1.7)}                           \\
Dropout  & 59.7    & 62.2 \textcolor{Green}{(+2.5)}                                 \\
Zoom in    & 64.3    & 65.6 \textcolor{Green}{(+1.3)}                                 \\
Zoom out  & 78.3    & 80.0 \textcolor{Green}{(+1.7)}                                 \\
Rotation      & 70.8    & 72.1 \textcolor{Green}{(+1.3)}                                 \\
Shearing      & 70.3    & 72.0 \textcolor{Green}{(+1.7)}                                 \\
Gaussian noise             & 65.1    & 67.2 \textcolor{Green}{(+2.1)}                                 \\
Gaussian blur             & 74.4    & 76.3 \textcolor{Green}{(+1.9)}                                \\
\bottomrule
\end{tabular}
\end{adjustbox}
\caption{Recall@1(\%) of input deformations with CARS196 trained models. Deformation details are presented in supplementary Section C.3.}
\label{table:input_robust}
\end{table}

\begin{figure}[t]
\centering
\begin{subfigure}[b]{0.45\columnwidth}
\centering
\includegraphics[width=1.0\linewidth]{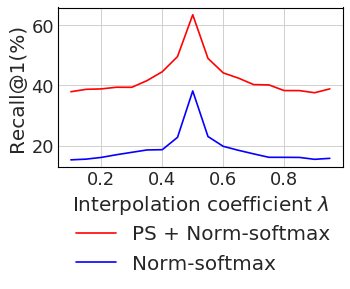}
\caption{Train set}
\label{fig:train_mixed}
\end{subfigure}
\begin{subfigure}[b]{0.45\columnwidth}
\centering
\includegraphics[width=1.01\linewidth]{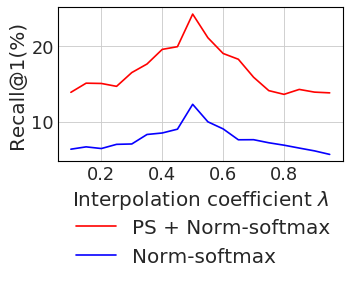}
\caption{Test set}
\label{fig:test_mixed}
\end{subfigure}
\caption{Recall@1(\%) of embedding deformations with trained networks on CARS196. For a gallery set, synthetic embeddings generated with $\lambda$ and original embeddings are used. For a query set, synthetic embeddings are used to find other synthetic embeddings generated with same manner.}
\label{fig:emb_robust}
\end{figure}

\begin{table*}[t]
\begin{adjustbox}{width=1.0\textwidth,center}
\begin{tabular}{>{\rowmac}l>{\rowmac}c>{\rowmac}c>{\rowmac}c>{\rowmac}c>{\rowmac}c>{\rowmac}c>{\rowmac}c>{\rowmac}c>{\rowmac}c<{\clearrow}}
\toprule
                  & \multicolumn{3}{c}{CUB200}                 & \multicolumn{3}{c}{CARS196}                & \multicolumn{3}{c}{SOP}                    \\ \cline{2-10} 
Loss              & P@1          & RP           & MAP@R        & P@1          & RP           & MAP@R        & P@1          & RP           & MAP@R        \\
\midrule
Norm-softmax      & 65.65 $\pm$ 0.30 & 35.99 $\pm$ 0.15 & 25.25 $\pm$ 0.13 & 83.16 $\pm$ 0.25 & 36.20 $\pm$ 0.26 & 26.00 $\pm$ 0.30 & 75.67 $\pm$ 0.17 & 50.01 $\pm$ 0.22 & 47.13 $\pm$ 0.22 \\
 \textit{PS} + Norm-softmax \setrow{\bfseries} & 69.19 $\pm$ 0.34 & 37.32 $\pm$ 0.29 & 26.40 $\pm$ 0.29 & 85.70 $\pm$ 0.24 & 38.33 $\pm$ 0.31 & 28.31 $\pm$ 0.32 & 76.73 $\pm$ 0.15 & 51.46 $\pm$ 0.21 & 48.52 $\pm$ 0.20 \\
CosFace           & 67.32 $\pm$ 0.32 & 37.49 $\pm$ 0.21 & 26.70 $\pm$ 0.23 & 85.52 $\pm$ 0.24 & 37.32 $\pm$ 0.28 & 27.57 $\pm$ 0.30 & 75.79 $\pm$ 0.14 & 49.77 $\pm$ 0.19 & 46.92 $\pm$ 0.19 \\
 \textit{PS} + CosFace  \setrow{\bfseries}    & 69.52 $\pm$ 0.26 & 37.99 $\pm$ 0.23 & 27.10 $\pm$ 0.23 & 85.58 $\pm$ 0.27 & 38.01 $\pm$ 0.19 & 27.89 $\pm$ 0.20 & 76.89 $\pm$ 0.20 & 51.60 $\pm$ 0.31 & 48.68 $\pm$ 0.33 \\
ArcFace           & 67.50 $\pm$ 0.25 & 37.31 $\pm$ 0.21 & 26.45 $\pm$ 0.20 & 85.44 $\pm$ 0.28 & 37.02 $\pm$ 0.29 & 27.22 $\pm$ 0.30 & 76.20 $\pm$ 0.27 & 50.27 $\pm$ 0.38 & 47.41 $\pm$ 0.40 \\
 \textit{PS} + ArcFace  \setrow{\bfseries}    & 68.79 $\pm$ 0.31 & 37.46 $\pm$ 0.26 & 26.79 $\pm$ 0.27 & 85.59 $\pm$ 0.25 & 38.31 $\pm$ 0.22 & 28.24 $\pm$ 0.20 & 77.21 $\pm$ 0.20 & 51.90 $\pm$ 0.23 & 49.02 $\pm$ 0.21 \\
SoftTriple        & 67.73 $\pm$ 0.39 & 37.34 $\pm$ 0.19 & 26.51 $\pm$ 0.20 & 84.49 $\pm$ 0.26 & 37.03 $\pm$ 0.21 & 28.07 $\pm$ 0.21 & 76.12 $\pm$ 0.17 & 50.21 $\pm$ 0.18 & 47.35 $\pm$ 0.19 \\
 \textit{PS} + SoftTriple \setrow{\bfseries}  & 68.26 $\pm$ 0.16 & 37.98 $\pm$ 0.21 & 27.02 $\pm$ 0.21 & 85.53 $\pm$ 0.12 & 38.40 $\pm$ 0.20 & 28.45 $\pm$ 0.19 & 77.59 $\pm$ 0.26 & 52.45 $\pm$ 0.21 & 49.53 $\pm$ 0.23 \\
 Proxy-NCA         & 65.69 $\pm$ 0.43 & 35.14 $\pm$ 0.26 & 24.21 $\pm$ 0.27 & 83.56 $\pm$ 0.27 & 35.62 $\pm$ 0.28 & 25.38 $\pm$ 0.31 & 75.89 $\pm$ 0.17 & 50.10 $\pm$ 0.22 & 47.22 $\pm$ 0.21 \\
 \textit{PS} + Proxy-NCA  \setrow{\bfseries}  & 66.02 $\pm$ 0.29 & 35.73 $\pm$ 0.24 & 24.84 $\pm$ 0.22 & 84.61 $\pm$ 0.19 & 36.39 $\pm$ 0.25 & 26.04 $\pm$ 0.27 & 76.78 $\pm$ 0.21 & 51.39 $\pm$ 0.27 & 48.44 $\pm$ 0.27 \\
Proxy-anchor      & 69.73 $\pm$ 0.31 & 38.23 $\pm$ 0.37 & 27.44 $\pm$ 0.35 & 86.20 $\pm$ 0.21 & 39.08 $\pm$ 0.31 & 29.37 $\pm$ 0.29 & 75.37 $\pm$ 0.15 & 50.19 $\pm$ 0.14 & 47.25 $\pm$ 0.15 \\
 \textit{PS} + Proxy-anchor \setrow{\bfseries} & 70.41 $\pm$ 0.36 & 38.82 $\pm$ 0.29 & 28.11 $\pm$ 0.29 & 86.90 $\pm$ 0.35 & 39.38 $\pm$ 0.27 & 29.71 $\pm$ 0.25 & 75.52 $\pm$ 0.21 & 50.45 $\pm$ 0.22 & 47.49 $\pm$ 0.20 \\
\bottomrule
\end{tabular}
\end{adjustbox}
\caption{\textbf{[MLRC evaluation]} Performance (\%) on the famous benchmarks of image retrieval task. We report the performance of concatenated 512-dim over 10 training runs. Bold numbers indicate the best score within the same loss and benchmark.}
\label{table:mlrc}
\end{table*}

\begin{table}[t]
\begin{adjustbox}{width=1.0\columnwidth,center}
\begin{tabular}{lcccc}
\toprule
\multirow{2}{*}{Regularizer} & \multicolumn{2}{c}{Softmax} & \multicolumn{2}{c}{Norm-softmax} \\ \cline{2-5}  
     & CARS196      & SOP          & CARS196         & SOP            \\ \midrule
Baseline        & 81.5         & 76.3         & 83.3            & 78             \\
Virtual Softmax & 77.3 \textcolor{Red}{(-4.2)}   & 76.2 \textcolor{Red}{(-0.1)}   &     -            &      -          \\
Input Mixup     & 81.1 \textcolor{Red}{(-0.4)}   & 77.0 \textcolor{Green}{(+0.7)}   & 82.2 \textcolor{Red}{(-1.1)}      & 78.2 \textcolor{Green}{(+0.2)}     \\
Manifold Mixup  & 81.6 \textcolor{Green}{(+0.1)}   & 77.5 \textcolor{Green}{(+1.2)}   & 83.6 \textcolor{Green}{(+0.3)}      & 78.4 \textcolor{Green}{(+0.4)}     \\
\textit{Proxy Synthesis} & \textbf{84.3 \textcolor{Green}{(+2.8)}}   & \textbf{78.1 \textcolor{Green}{(+1.8)}}   & \textbf{84.7 \textcolor{Green}{(+1.4)}}      & \textbf{79.6 \textcolor{Green}{(+1.6)}}    \\
\bottomrule
\end{tabular}
\end{adjustbox}
\caption{Recall@1(\%) comparison with other regularizers in image retrieval task.}
\label{table:regularizer}
\end{table}

\subsection{Robustness to Deformation}
\label{sec:deformation}

\subsubsection{Input Deformation:}
To further evaluate the quality of representations learned with \textit{Proxy Synthesis}, we perform a deformation test on the input data with trained networks.  
In Table~\ref{table:input_robust}, we evaluate the test data with several input deformations that are not used in training.
A better-generalized model should be more robust to a large variety of input deformations.
The result indicates that the network trained using \textit{Proxy Synthesis} demonstrates significantly improved performance to input deformations.

\subsubsection{Embedding Deformation:}
To see the robustness on embedding deformation of trained networks, we evaluate performance with synthetic embeddings.
Figure~\ref{fig:emb_robust} depicts a network trained with Norm-softmax loss struggling with low performance on both the train and test set.
In contrast, a network trained with \textit{Proxy Synthesis} performs almost twice as well when compared with Norm-softmax loss on both the train and test set.
This demonstrates that \textit{Proxy Synthesis} provides more robust embedding features, which also leads to robustness on \textit{unseen} classes.
Besides, the patterns of performance are similar to those discussed in Section~\ref{sec:hardness}.
When $\lambda$ is close to 0 and 1, the performance is low because of hard synthetics, and when $\lambda$ is close to 0.5, the performance is high because of relatively easy synthetics.
Additional experiments are in the supplementary Section D.3.

\subsection{Comparison with Other Regularizers}
\label{sec:comparison}

Further, we compare the proposed method with other regularizers, including Virtual Softmax, Input Mixup, and Manifold Mixup in the image retrieval task.
Note that Virtual Softmax is not applicable to Norm-softmax loss because $W_{virt}$ will always be constant $1$.
As presented in Table~\ref{table:regularizer}, Virtual Softmax degrades the performance of all cases with a margin of average -2.15\%.
Input Mixup degrades the performance on CARS196 with an average margin -0.75\% and improves the performance on SOP with an average margin +0.45\%.
Manifold Mixup increases the performance of all cases with an average margin +0.5\%.
This illustrates that although these techniques are powerful for generalizing \textit{seen} classes, such as classification tasks, they lack discriminative ability on \textit{unseen} classes.
On the other hand, \textit{Proxy Synthesis} improves performance for all cases with a large margin of average +1.9\% and achieves the best performance among all.
Further analysis, including hyper-parameter search for Mixup and experiments in the classification task, is presented in the supplementary material Section D.4.

\begin{table}[t]
\begin{adjustbox}{width=1.0\columnwidth,center}
\begin{tabular}{>{\rowmac}l>{\rowmac}c>{\rowmac}c>{\rowmac}c>{\rowmac}c<{\clearrow}}
\toprule
Method            & CUB200  & CARS196 & SOP  & In-Shop \\
\midrule
Softmax           & 64.2        & 81.5        & 76.3        & 90.4        \\
\textit{PS} + Softmax \setrow{\bfseries}     & 64.9 \textcolor{Green}{(+0.7)} & 84.3 \textcolor{Green}{(+2.8)} & 77.6 \textcolor{Green}{(+1.3)} & 90.9 \textcolor{Green}{(+0.5)} \\
Norm-softmax      & 64.9        & 83.3        & 78.6        & 90.4        \\
\textit{PS} + Norm-softmax \setrow{\bfseries} & 66.0 \textcolor{Green}{(+1.1)} & 84.7 \textcolor{Green}{(+1.4)} & 79.6 \textcolor{Green}{(+1.0)} & 91.5 \textcolor{Green}{(+1.1)} \\
SphereFace        & 65.4        & 83.6        & 78.9        & 90.3        \\
\textit{PS} + SphereFace \setrow{\bfseries}  & 66.6 \textcolor{Green}{(+1.2)} & 85.1 \textcolor{Green}{(+1.5)} & 79.4 \textcolor{Green}{(+0.5)} & 91.6 \textcolor{Green}{(+1.3)} \\
CosFace           & 65.7        & 83.6        & 78.6        & 90.7        \\
\textit{PS} + CosFace  \setrow{\bfseries}    & 66.6 \textcolor{Green}{(+0.9)} & 84.6 \textcolor{Green}{(+1.0)} & 79.3 \textcolor{Green}{(+0.7)} & 91.4 \textcolor{Green}{(+0.7)} \\
ArcFace           & 66.1        & 83.7        & 78.8        & 91.0        \\
 \textit{PS} + ArcFace  \setrow{\bfseries}    & 66.8 \textcolor{Green}{(+0.7)} & 84.7 \textcolor{Green}{(+1.0)} & 79.7 \textcolor{Green}{(+0.9)} & 91.7 \textcolor{Green}{(+0.7)} \\
Proxy-NCA         & 65.1        & 83.7        & 78.1        & 90.0        \\
 \textit{PS} + Proxy-NCA \setrow{\bfseries}  & 66.4 \textcolor{Green}{(+1.3)} & 84.5 \textcolor{Green}{(+0.8)} & 79.1 \textcolor{Green}{(+1.0)} & 91.4 \textcolor{Green}{(+1.4)} \\
SoftTriple        & 65.4        & 84.5        & 78.3        & 91.1        \\
 \textit{PS} + SoftTriple \setrow{\bfseries}  & 66.6 \textcolor{Green}{(+1.2)} & 85.3 \textcolor{Green}{(+0.8)} & 79.5 \textcolor{Green}{(+1.2)} & 91.8 \textcolor{Green}{(+0.7)} \\
Proxy-anchor$^\dagger$      & 68.4        & 86.1        & 79.1        & 91.5        \\
 \textit{PS} + Proxy-anchor$^\dagger$ \setrow{\bfseries} & 69.2 \textcolor{Green}{(+0.8)} & 86.9 \textcolor{Green}{(+0.8)} & 79.8 \textcolor{Green}{(+0.7)} & 91.9 \textcolor{Green}{(+0.4)} \\
\bottomrule
\end{tabular}
\end{adjustbox}
\caption{\textbf{[Conventional evaluation]} Recall@1 (\%) on the famous benchmarks of image retrieval task. Bold numbers indicate the best score within the same loss and benchmark. $^\dagger$ denotes exceptional experimental settings as described in the supplementary Section C.2.}
\label{table:sota}
\end{table}

\subsection{Comparison with State-of-the-Art}
\label{sec:sota}
Finally, we compare the performance of our proposed method with state-of-the-art losses in two ways: \textit{conventional} and \textit{MLRC evaluation}.
In \textit{conventional evaluation}, the combinations of \textit{Proxy Synthesis} with proxy-based losses improve performance by a large margin in every benchmark as presented in Table~\ref{table:sota}.
For fine-grained datasets with few categories such as CUB200 and CARS196, the performance gain ranges between a minimum of +0.7\% and a maximum of +2.8\%, and the average improvement is +1.1\%.
For large-scale datasets with numerous categories such as SOP and In-Shop, the performance gain ranges between a minimum of +0.4\% and a maximum of +1.4\%, and the average improvement is +0.9\%.
Even in the specifically designed \textit{MLRC evaluation}, Table~\ref{table:mlrc} shows that \textit{Proxy Synthesis} enhances performance for every metric and benchmark.
Extended comparisons with Recall@k for \textit{conventional evaluation} and performance of seperated 128-dim for \textit{MLRC evaluation} are presented in the supplementary Section D.6.

\section{Conclusion}
\label{sec:conclusion}

In this paper, we have proposed a novel regularizer called \textit{Proxy Synthesis} for proxy-based losses that exploits  synthetic classes for stronger generalization.
Such effect is achieved by deriving class relations and smoothened decision boundaries.
The proposed method provides a significant performance boost for all proxy-based losses and achieves state-of-the-art performance in image retrieval tasks.

\paragraph{Acknowledgement}
We would like to thank Yoonjae Cho, Hyong-Keun Kook, Sanghyuk Park, Minchul Shin and Tae Kwan Lee from NAVER/LINE Vision team for helpful comments and feedback.

\bibliography{egbib}

\clearpage



\section*{Supplementary material}

# Proxy Synthesis: Learning with Synthetic Class for Deep Metric Learning
## - *Supplementary Material* -

**Geonmo Gu**[*1], **Byungsoo Ko**[*1], **Han-Gyu Kim** [2]

[1] NAVER/LINE Vision, [2] NAVER Clova Speech
korgm403@gmail.com, kobiso62@gmail.com, hangyu.kim@navercorp.com
[github.com/navervision/proxy-synthesis](github.com/navervision/proxy-synthesis)

## Contents

---

[*]Authors contributed equally.

# A  Proxy-based Losses

In this section, we describe each proxy-based loss, including softmax and its variants used in our study. We follow the same notation from Section 3 in the main paper.

**Proxy-NCA:** Proxy-NCA loss (Movshovitz-Attias et al. 2017) is the first proxy-based loss in metric learning to resolve sampling issues of pair-based losses. It proposes the concept of *proxy*, which approximates the original data points so that we do not need to sample informative pairs but can explicitly compute the loss with proxies. Proxy-NCA loss is defined as:

$$\mathcal{L}_{ProxyNCA}(X, P) = -\frac{1}{|X|} \sum_{x \in X} \log \frac{e^{-d(x,p^+)}}{\sum_{q \in P^-} e^{-d(x,q)}}, \tag{i}$$

where $d(a,b) = \|a - b\|_2$ is the Euclidean distance, $p^+$ is positive proxy and $P^-$ denotes the set of negative proxies.

**SoftTriple:** SoftTriple loss (Qian et al. 2019) extends softmax loss with multiple proxies (centers) for each class. It assigns each image to one of the sub-proxies within the class, which has the maximum similarity. We define the relaxed similarity between the sample $x$ and the proxy $p$ as:

$$R(x, p) = \sum_k \frac{e^{s(x,p_k)/\gamma}}{\sum_k e^{s(x,p_k)/\gamma}} s(x, p_k), \tag{ii}$$

where $s(a, b) = a^T b$ denotes the cosine similarity between feature $a$ and $b$, $p_k$ is a $k$-th sub-proxy of the class, and $\gamma > 0$ is a scaling factor. Then, SoftTriple loss is presented as:

$$\mathcal{L}_{SoftTriple}(X, P) = -\frac{1}{|X|} \sum_{x \in X} \log \left\{ \frac{e^{\epsilon R(x,p^+) - \delta}}{e^{\epsilon R(x,p^+) - \delta} + \sum_{q \in P^-} e^{\epsilon R(x,q)}} \right\}, \tag{iii}$$

where $\epsilon > 0$ is a scaling factor and $\delta > 0$ is a margin.

**Proxy-anchor:** Proxy-anchor loss (Kim et al. 2020) is one of the most recent proxy-based losses in the metric learning task. Unlike the existing proxy-based losses, it exploits each proxy as an anchor and considers its relations of all data in a batch. For each proxy $p$, let $X_p^+$ and $X_p^-$ as the set of positive and negative embedding vectors of $p$, respectively, where $X$ is divided into two sets of $X_p^+$ and $X_p^-$. Proxy-anchor loss can be formulated as:

$$\mathcal{L}_{ProxyAnchor}(X, P) = \frac{1}{|P^+|} \sum_{p \in P^+} \log \left\{ 1 + \sum_{x \in X_p^+} e^{-\gamma(s(x,p) - \delta)} \right\} + \frac{1}{|P|} \sum_{p \in P} \log \left\{ 1 + \sum_{x \in X_p^-} e^{\gamma(s(x,p) + \delta)} \right\}, \tag{iv}$$

where $\gamma > 0$ is a scaling factor, and $\delta > 0$ is a margin.

**Softmax:** As formulated in Equation 2 in the main paper Section 3.1, Softmax loss is presented as follows:

$$\mathcal{L}_{Softmax}(X) = -\frac{1}{|X|} \sum_{i=1}^{|X|} \log \frac{e^{W_{y_i}^T x_i}}{\sum_{j=1}^{C} e^{W_j^T x_i}}, \tag{v}$$

When we transform the logit (Pereyra et al. 2017) as $W_j^T x_i = \|W_j\| \|x_i\| \cos \theta_j$, where $\theta_j$ is an angle between the weight $W_j$ and the feature $x_i$, the softmax loss can be rewritten with proxy-wise form as:

$$\mathcal{L}_{Softmax}(X, P) = -\frac{1}{|X|} \sum_{x \in X} \log \left\{ \frac{e^{\|p^+\| \|x\| s(x,p^+)}}{e^{\|p^+\| \|x\| s(x,p^+)} + \sum_{q \in P^-} e^{\|q\| \|x\| s(x,q)}} \right\}. \tag{vi}$$

**Normalized Softmax:** As defined in Equation 3 in Section 3.1, the normalized softmax (Norm-softmax) loss, which normalizes the weights and feature vectors, is presented as follows:

$$\mathcal{L}_{Norm}(X, P) = -\frac{1}{|X|} \sum_{x \in X} \log \left\{ \frac{e^{\gamma s(x,p^+)}}{e^{\gamma s(x,p^+)} + \sum_{q \in P^-} e^{\gamma s(x,q)}} \right\}. \tag{vii}$$

```
Algorithm 1: Pseudo-code of Proxy Synthesis
1  def compute_loss(batch_size, embeddings, proxies, α, μ):
2      if proxy_synthesis then
3          λ = beta_distribution(α, α)
4          n_syn = μ×batch_size
5          x, x' = sample_pairs(embeddings, n_syn)    // sample n_syn embeddings from different classes
6          p, p' = sample_pairs(proxies, n_syn)       // sample n_syn proxies from different classes
7          x̃ = λ × x + (1 − λ) × x'                   // generate synthetic embeddings by linear interpolation
8          p̃ = λ × p + (1 − λ) × p'                   // generate synthetic proxies by linear interpolation
9          embeddings = concatenate(embeddings, x̃)    // merge synthetic and original embeddings
10         proxies = concatenate(proxies, p̃)          // merge synthetic and original proxies
11     loss = compute_proxy_based_loss(embeddings, proxies)
12     return loss
```

**Softmax Variants:** To directly enhance the feature discrimination, several softmax variant losses have been proposed by adding margin penalties in SphereFace (Liu et al. 2017), ArcFace (Deng et al. 2019), and CosFace (Wang et al. 2018). An integrated framework of these three losses with $m_1$, $m_2$, and $m_3$ hyper-parameters, respectively, is formulated as:

$$\mathcal{L}_{Variants}(X, P) = -\frac{1}{|X|} \sum_{x \in X} \log \left\{ \frac{e^{\gamma(\cos(m_1 \arccos s(x,p^+) + m_2) - m_3)}}{e^{\gamma(\cos(m_1 \arccos s(x,p^+) + m_2) - m_3)} + \sum_{q \in P^-} e^{\gamma s(x,q)}} \right\}. \tag{viii}$$

Note that this includes the Norm-softmax loss when $m_1 = 1.0, m_2 = 0.0$, and $m_3 = 0.0$.

## B  Details of Proxy Synthesis

### B.1  Proxy Synthesis Algorithm

We present a code-level description of *Proxy Synthesis* in Algorithm 1. In the method arguments, *embeddings* and *proxies* are list of original embeddings and proxies. $\alpha$ is a shape parameter for beta distribution, and $\mu$ is a generation ratio by batch size as defined in Section 3.2 from the main paper. First, *Proxy Synthesis* samples pairs of embeddings and proxies from different classes, respectively. And we generate synthetic embeddings and proxies by linear interpolation. Then, we merge synthetic and original embeddings and proxies to use in loss computation. Note that *Proxy Synthesis* is extremely easy and simple to implement with only several lines of code. Furthermore, it does not need to revise any code of proxy-based loss and can be used in a plug-and-play manner.

### B.2  Induction for Gradient Analysis

We provide a detailed induction of gradient analysis, which is discussed in *Learning with Class Relations* from Section 3.3. We first show the induction of gradient analysis for the loss of softmax function, and then present the induction of gradient analysis when *Proxy Synthesis* is applied.

**Softmax:** The loss of softmax function on $(x, p_i)$, where $x$ is an anchor embedding of input, and $p_i$ is the corresponding positive proxy, can be written as follows,

$$\begin{aligned}
\mathcal{L}_i &= \mathcal{L}_{Softmax}(x, p_i) \\
&= -\log \frac{e^{S(x,p_i)}}{e^{S(x,p_i)} + \sum_{q \in P^-} e^{S(x,q)}} \\
&= -\log \frac{E(p_i)}{E(p_i) + \sum_{q \in P^-} E(q)},
\end{aligned} \tag{ix}$$

where $E(p) = e^{S(x,p)}$, $S(x,p) = s(x,p) \parallel x \parallel \parallel p \parallel = x^T p$ and $P^- = P \setminus \{p_i\}$. Then, the gradient over $S(x, p_i)$ is

$$
\begin{aligned}
\frac{\partial \mathcal{L}_i}{\partial S(x, p_i)} &= \frac{\partial \mathcal{L}_i}{\partial E(p_i)} \frac{\partial E(p_i)}{\partial S(x, p_i)} \\
&= -\frac{E(p_i) + \sum_{q \in P^-} E(q)}{E(p_i)} \frac{\left(E(p_i) + \sum_{q \in P^-} E(q)\right) - E(p_i)}{\left(E(p_i) + \sum_{q \in P^-} E(q)\right)^2} E(p_i) \\
&= -\frac{\sum_{q \in P^-} E(q)}{E(p_i) + \sum_{q \in P^-} E(q)} \\
&= \frac{E(p_i)}{\sum_{q \in P} E(q)} - 1. \quad (x)
\end{aligned}
$$

Besides the gradient mentioned in the original paper, we have also inducted the gradient over $S(x, p_k)$ where $p_k \neq p_i$ additionally. The inducted gradient is as follows,

$$
\begin{aligned}
\frac{\partial \mathcal{L}_i}{\partial S(x, p_k)} &= \frac{\partial \mathcal{L}_i}{\partial E(p_k)} \frac{\partial E(p_k)}{\partial S(x, p_k)} \\
&= -\frac{E(p_i) + \sum_{q \in P^-} E(q)}{E(p_i)} \frac{0 - E(p_i)}{\left(E(p_i) + \sum_{q \in P^-} E(q)\right)^2} E(p_k) \\
&= \frac{E(p_k)}{\sum_{q \in P} E(q)}. \quad (xi)
\end{aligned}
$$

**Proxy Synthesis:** In order to show the major difference between ordinary sofmax and *Proxy Synthesis*, we assume that the positive proxy $p_i$ of input $x$ is used for generating synthesized proxy $\tilde{p}$ with $p_j$ as $\tilde{p} = \lambda p_i + \lambda' p_j$, where $\lambda' = 1 - \lambda$. Then the loss function in Equation ix should be re-written as follows,

$$
\begin{aligned}
\mathcal{L}_i &= \mathcal{L}_{Softmax}(x, p_i) \\
&= -\log \frac{e^{S(x,p_i)}}{e^{S(x,\tilde{p})} + e^{S(x,p_i)} + e^{S(x,p_j)} + \sum_{q \in P^-} e^{S(x,q)}} \\
&= -\log \frac{E(p_i)}{E(\tilde{p}) + E(p_i) + E(p_j) + \sum_{q \in P^-} E(q)} \\
&= -\log \frac{E(p_i)}{E(p_i)^\lambda E(p_j)^{\lambda'} + E(p_i) + E(p_j) + \sum_{q \in P^-} E(q)},
\end{aligned}
$$

In this induction, the gradients over three parameters are interested, which are $S(x, p_i)$, $S(x, p_j)$ and $S(x, p_k)$, where $p_k \neq p_i, p_j$. The induction of each gradient is written as follows.

Case 1. the gradient over $S(x, p_i)$:

$$
\begin{aligned}
\frac{\partial \mathcal{L}_i}{\partial S(x, p_i)} &= \frac{\partial \mathcal{L}_i}{\partial E(p_i)} \frac{\partial E(p_i)}{\partial S(x, p_i)} \\
&= -\frac{E(p_i)^\lambda E(p_j)^{\lambda'} + E(p_i) + E(p_j) + \sum_{q \in P^-} E(q)}{E(p_i)} \\
&\quad \times \frac{\left(E(p_i)^\lambda E(p_j)^{\lambda'} + E(p_i) + E(p_j) + \sum_{q \in P^-} E(q)\right) - E(p_i)\left(\lambda E(p_i)^{\lambda-1} E(p_j)^{\lambda'} + 1\right)}{\left(E(p_i)^\lambda E(p_j)^{\lambda'} + E(p_i) + E(p_j) + \sum_{q \in P^-} E(q)\right)^2} E(p_i) \\
&= -\frac{\lambda' E(p_i)^\lambda E(p_j)^{\lambda'} + E(p_j) + \sum_{q \in P^-} E(q)}{E(p_i)^\lambda E(p_j)^{\lambda'} + E(p_i) + E(p_j) + \sum_{q \in P^-} E(q)} \\
&= \frac{\lambda E(\tilde{p}) + E(p_i)}{E(\tilde{p}) + \sum_{q \in P} E(q)} - 1. \quad (xii)
\end{aligned}
$$

Case 2. the gradient over $S(x, p_j)$:

$$
\begin{aligned}
\frac{\partial \mathcal{L}_i}{\partial S(x, p_j)} &= \frac{\partial \mathcal{L}_i}{\partial E(p_j)} \frac{\partial E(p_j)}{\partial S(x, p_j)} \\
&= -\frac{E(p_i)^\lambda E(p_j)^{\lambda'} + E(p_i) + E(p_j) + \sum_{q \in P^-} E(q)}{E(p_i)} \\
&\quad \times \frac{0 - E(p_i)\left(\lambda' E(p_i)^\lambda E(p_j)^{\lambda'-1} + 1\right)}{\left(E(p_i)^\lambda E(p_j)^{\lambda'} + E(p_i) + E(p_j) + \sum_{q \in P^-} E(q)\right)^2} E(p_j) \\
&= \frac{\lambda' E(p_i)^\lambda E(p_j)^{\lambda'} + E(p_j)}{E(p_i)^\lambda E(p_j)^{\lambda'} + E(p_i) + E(p_j) + \sum_{q \in P^-} E(q)} \\
&= \frac{\lambda' E(\tilde{p}) + E(p_j)}{E(\tilde{p}) + \sum_{q \in P} E(q)}.
\end{aligned} \quad \text{(xiii)}
$$

Case 3. the gradient over $S(x, p_k)$:

$$
\begin{aligned}
\frac{\partial \mathcal{L}_i}{\partial S(x, p_k)} &= \frac{\partial \mathcal{L}_i}{\partial E(p_k)} \frac{\partial E(p_k)}{\partial S(x, p_k)} \\
&= -\frac{E(p_i)^\lambda E(p_j)^{\lambda'} + E(p_i) + E(p_j) + \sum_{q \in P^-} E(q)}{E(p_i)} \\
&\quad \times \frac{0 - E(p_i)}{\left(E(p_i)^\lambda E(p_j)^{\lambda'} + E(p_i) + E(p_j) + \sum_{q \in P^-} E(q)\right)^2} E(p_k) \\
&= \frac{E(p_k)}{E(p_i)^\lambda E(p_j)^{\lambda'} + E(p_i) + E(p_j) + \sum_{q \in P^-} E(q)} \\
&= \frac{E(p_k)}{E(\tilde{p}) + \sum_{q \in P} E(q)}.
\end{aligned} \quad \text{(xiv)}
$$

### B.3 Proof of Smooth Decision Boundary

We present a detailed induction for proof of smooth decision boundary, which is discussed in Section 3.3. We show the proof by comparing gradient descent of softmax and *Proxy Synthesis*.

**Gradient Descent of Softmax:** The induction of the gradient of loss function $\mathcal{L}_i$ over embedding vector $x$ in Equation ix is as follows:

$$
\begin{aligned}
\frac{\partial \mathcal{L}_i}{\partial x} &= -\frac{E(p_i) + \sum_{q \in P^-} E(q)}{E(p_i)} \\
&\quad \times \frac{p_i E(p_i)\left(E(p_i) + \sum_{q \in P^-} E(q)\right) - E(p_i)\left(p_i E(p_i) + \sum_{q \in P^-} q E(q)\right)}{\left(E(p_i) + \sum_{q \in P^-} E(q)\right)^2} \\
&= -\frac{p_i \sum_{q \in P^-} E(q) - \sum_{q \in P^-} q E(q)}{\sum_{q \in P} E(q)} \\
&= -\frac{p_i \sum_{q \in P^-} E(q) - \sum_{p_k \in P^-} p_k E(p_k)}{\sum_{q \in P} E(q)} \\
&= \left(\frac{E(p_i)}{\sum_{q \in P} E(q)} - 1\right) p_i + \sum_{p_k \in P^-} \frac{E(p_k)}{\sum_{q \in P} E(q)} p_k, \\
&= \tau_i p_i + \sum_{p_k \in P^-} \tau_k p_k,
\end{aligned} \quad \text{(xv)}
$$

where,

$$\tau_i = \frac{E(p_i)}{\sum_{q \in P} E(q)} - 1,$$

$$\tau_k = \frac{E(p_k)}{\sum_{q \in P} E(q)} \quad (p_k \neq p_i). \tag{xvi}$$

It is obvious that $\tau_i < 0$ and $\tau_k > 0$. Considering the parameter update is performed by $x = x - \eta \frac{\partial \mathcal{L}_i}{\partial x}$, where $\eta$ is a learning rate, the gradient descent always forces $x$ to be closer to $p_i$ and to be distant from other proxies $p_k$.

**Gradient Descent of Proxy Synthesis:** In order to describe major difference of *Proxy Synthesis* and ordinary softmax, we consider softmax loss for synthesized pair $\tilde{x} = \lambda x + \lambda' x', \tilde{p} = \lambda p_i + \lambda' p_j$, where $(x, p_i)$ and $(x', p_j)$ are pairs of an embedding and a corresponding proxy,

$$\begin{aligned}
\tilde{\mathcal{L}} &= \mathcal{L}_{Softmax}(\tilde{x}, \tilde{p}) \\
&= -\log \frac{e^{S(\tilde{x}, \tilde{p})}}{e^{S(\tilde{x}, \tilde{p})} + e^{S(\tilde{x}, p_i)} + e^{S(\tilde{x}, p_j)} + \sum_{q \in P^-} e^{S(\tilde{x}, q)}}, \\
&= -\log \frac{E(\tilde{p})}{E(\tilde{p}) + E(p_i) + E(p_j) + \sum_{q \in P^-} E(q)},
\end{aligned} \tag{xvii}$$

where $E(p) = e^{S(\tilde{x}, p)}$ and $P^- = P \setminus \{p_i, p_j\}$. It should be noted that $\tilde{p} \notin P$. Since we suggest to sample $\lambda$ from $Beta(\alpha, \alpha)$ with small $\alpha$, $\tilde{p}$ has high chance to be generated either close to $p_i$ with $\lambda >> 0.5$ or close to $p_j$ with $\lambda << 0.5$. As the proofs for both cases are equivalent, we assume the first case: $\tilde{x}$ is much closer to $x$ than $x'$. We consider gradient over $x$ because the loss will affect the closer embedding vector more than the other one. The desired gradient is inducted as follows,

$$\begin{aligned}
\frac{\partial \tilde{\mathcal{L}}}{\partial x} &= -\frac{E(\tilde{p}) + E(p_i) + E(p_j) + \sum_{q \in P^-} E(q)}{E(\tilde{p})} \\
&\quad \times \frac{\lambda \tilde{p} E(\tilde{p}) \big( E(\tilde{p}) + E(p_i) + E(p_j) + \sum_{q \in P^-} E(q) \big) - E(\tilde{p}) \big( \lambda \tilde{p} E(\tilde{p}) + \lambda p_i E(p_i) + \lambda p_j E(p_j) + \sum_{q \in P^-} \lambda q E(q) \big)}{\big( E(\tilde{p}) + E(p_i) + E(p_j) + \sum_{q \in P^-} E(q) \big)^2} \\
&= -\lambda \frac{(\tilde{p} - p_i) E(p_i) + (\tilde{p} - p_j) E(p_j) + \sum_{q \in P^-} (\tilde{p} - q) E(q)}{E(\tilde{p}) + E(p_i) + E(p_j) + \sum_{q \in P^-} E(q)} \\
&= -\lambda \frac{\sum_{q \in P} (\tilde{p} - q) E(q)}{E(\tilde{p}) + \sum_{q \in P} E(q)}.
\end{aligned} \tag{xviii}$$

As mentioned above, $\tilde{x}$ and $\tilde{p}$ are much closer to $x$ and $p_i$ respectively, compared to other class. Therefore, $E(\tilde{p}), E(p_i) >> E(q)$ ($\forall q \neq \tilde{p}, p_i$). Although $|\tilde{p} - p_i| < |\tilde{p} - q|$ ($\forall q \neq \tilde{p}, p_i$), as the difference of exponential function is much larger than the difference of linear function, $|(\tilde{p} - p_i) E(p_i)| >> |(\tilde{p} - q) E(q)|$ ($\forall q \neq \tilde{p}, p_i$), implying that $(\tilde{p} - p_i) E(p_i)$ is the dominant term of $\sum_{q \in P} (\tilde{p} - q) E(q)$ and $E(p_i)$ is the dominant term of $\sum_{q \in P} E(q)$ in Equation xviii. With such conclusion, the above gradient may be approximated as follows,

$$\begin{aligned}
\frac{\partial \tilde{\mathcal{L}}}{\partial x} &\approx -\lambda \frac{(\tilde{p} - p_i) E(p_i)}{E(\tilde{p}) + E(p_i)} \\
&= \lambda \lambda' \frac{(p_i - p_j) E(p_i)}{E(\tilde{p}) + E(p_i)} \\
&= \tau_i' p_i + \tau_j' p_j,
\end{aligned} \tag{xix}$$

where,

$$\begin{aligned}
\tau_i' &= \frac{\lambda \lambda' E(p_i)}{E(\tilde{p}) + E(p_i)}, \\
\tau_j' &= -\frac{\lambda \lambda' E(p_i)}{E(\tilde{p}) + E(p_i)}.
\end{aligned} \tag{xx}$$

It is obvious that $\tau_i' > 0$, implying that by adopting *Proxy Synthesis*, softmax loss for synthesized pair provides gradient which leads embedding vector not too close to the corresponding proxy $p_i$; it is also obvious that $\tau_j' < 0$, implying that softmax loss for synthesized pair provides gradient which makes embedding vector not too distant from the competing proxy $p_j$. In such a manner, *Proxy Synthesis* prevents embedding vectors lying too close to proxies, which finally leads to the smooth decision boundary.

# C  Details of Experimental Setting

## C.1  Datasets

We employ four widely used benchmarks in metric learning for the evaluation: CUB-200-2011 (CUB200) (Wah et al. 2011), CARS196 (Krause et al. 2013), Standford Online Products (SOP) (Oh Song et al. 2016), and In-Shop Clothes (In-Shop) (Liu et al. 2016).

- For CUB200, we use 5,864 images of the first 100 classes for training and the remaining 5,924 images of the 100 other classes for testing.
- For CARS196, we use the first 8,054 images of 98 classes for training and the remaining 8,131 images of the 98 other classes for testing.
- For SOP, we follow the train and test splits described in (Oh Song et al. 2016), where the first 59,551 images of 11,318 classes are used for training, and the remaining 60,502 images of 11,316 classes are used for testing.
- For In-Shop, we follow the setting described in (Liu et al. 2016) using 25,882 images of the first 3,997 classes for training and 28,760 images of the remaining classes for testing; the test set is further partitioned into a query set with 14,218 images of 3,985 classes and a gallery set with 12,612 images of 3,985 classes.

Note that we do not use the bounding box information for CUB200 and CARS196.

## C.2  Implementation

Throughout the experiments, we use the PyTorch (Paszke et al. 2019) framework on a Tesla P40 GPU with 24GB memory. In the "conventional evaluation", we follow the widely used training and evaluation procedure from (Oh Song et al. 2016; Sohn 2016; Kim et al. 2020; Wang et al. 2019a). In the "metric learning reality check (MLRC) evaluation", we follow the specifically designed procedure in terms of fairness from (Musgrave, Belongie, and Lim 2020).

**Conventional Evaluation:**  The input images are augmented by random cropping and horizontal flipping in the training phase, whereas they are center-cropped in the test phase. The size of the cropped images is $224 \times 224$. For the backbone network, we use ImageNet (Deng et al. 2009) pre-trained Inception network with batch normalization (Ioffe and Szegedy 2015) (BN-Inception). We freeze batch normalization for CUB200 and CARS196 because the images obtained from them are similar to those obtained from ImageNet (Deng et al. 2009), whereas we keep BN training on SOP and In-Shop by following (Qian et al. 2019; Wang et al. 2019a; Kim et al. 2020). With a global average pooling followed by a fully connected layer for dimensional reduction, we set the embedding feature's output dimension to 512. We set the learning rate of $10^{-4}$ for CUB200 and CARS196, and $10^{-3}$ for SOP and In-Shop using the Adam optimizer (Kingma and Ba 2014) and a batch size of 128. The learning rate is decayed by a factor of 0.1 at the 50th epoch for CARS196, 60th epoch for In-Shop, and 20th epoch for CUB200 and SOP. We use $\alpha = 0.4$ and $\mu = 1.0$ for *Proxy Synthesis* in every experiment, except that we search for the best values in the experiment of comparison with state-of-the-art. For Proxy-anchor loss, we follow the experimental setting of (Kim et al. 2020), which includes batch size of 180, AdamW optimizer (Loshchilov and Hutter 2017), parameter warm-up, and combination of average and max pooling layers. Considering that papers of softmax variants (Liu et al. 2017; Wang et al. 2018; Deng et al. 2019) do not include hyper-parameters for the aforementioned benchmarks, we search for the best hyper-parameters. In Eq 6, $\gamma = 30.0$, $m_1 = 1.05$, $m_2 = 0.0$, and $m_3 = 0.0$ is used for SphereFace (Liu et al. 2017), $\gamma = 23.0$, $m_1 = 1.0$, $m_2 = 0.0$, and $m_3 = 0.1$ is used for CosFace (Wang et al. 2018), and $\gamma = 23.0$, $m_1 = 1.0$, $m_2 = 0.1$, and $m_3 = 0.0$ is used for ArcFace (Deng et al. 2019). For the rest of the losses, we follow the same hyper-parameters as specified in each paper.

**MLRC Evaluation:**  In the training phase, we first resize each image to make its shorter side has length 256, then perform a random crop to have a size between 40 and 256, and aspect ratio between 3/4 and 4/3. This image is then resized to $227 \times 227$ and flipped horizontally with a 50% probability. During the test phase, images are resized 256 and then center cropped to 227. For the backbone network, we use an ImageNet pre-trained BN-Inception network with an output embedding size of 128. All networks are trained with a batch size of 32, RMSprop optimizer, and learning rate of $10^{-6}$. In order to find the best hyperparameters for loss functions, we conduct 50 experiments of hyper-parameter search where each experiment consists of 4-fold cross-validation. We perform 10 training using the best hyper-parameters and report the average and confidence intervals across these experiments. We report separated (128-dim) and concatenated (512-dim) accuracy, which is concatenated and $l_2$-normalized of 128-dim embeddings of the 4 models.

| | Time(*ms*) | | Memory |
| $\mu$ | Generation | Loss | # of Features |
| --- | --- | --- | --- |
| 0.0 | - | 0.558 | $N$ |
| 0.2 | 0.232 | 0.872 | $(1+0.2) \times N$ |
| 0.5 | 0.233 | 0.873 | $(1+0.5) \times N$ |
| 1.0 | 0.435 | 1.090 | $(1+1.0) \times N$ |
| 1.5 | 0.449 | 1.300 | $(1+1.5) \times N$ |
| 2.0 | 0.467 | 1.530 | $(1+2.0) \times N$ |

Table A: Computation time and memory based on various values $\mu$. With batch size $N = 128$, time required for generating synthetic embeddings and proxies (Generation), and computing loss (Loss) is measured, and number of features for both embeddings and proxies are calculated. *PS + Norm-softmax* loss is used on CARS196 with $\mu = 0.0$ as baseline.

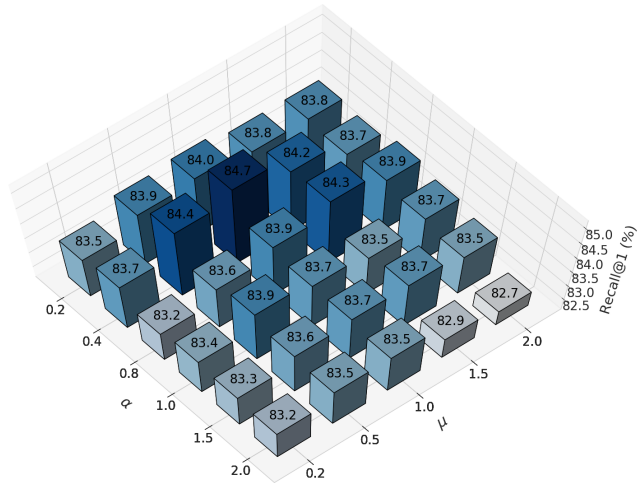

Figure A: Recall@1(%) versus hyper-parameters $\alpha$ and $\mu$, where network is trained with *PS + Norm-softmax* loss on CARS196.

## C.3 Input Deformation

For the input deformation experiment described in Section 4.4, we use the *imgaug* (Jung et al. 2020) python library for image deformation.

- In the *cutout* experiment, each input image is randomly filled by two grayish pixels that are 20% of the image size.
- For the *dropout* experiment, a value $p$ from the range of $0 \leq p \leq 0.2$ is sampled; thereafter, $p \times 100\%$ of all pixels is dropped from each image.
- For the *zoom-in* and *zoom-out* experiments, each input image is transformed by zoom-in and zoom-out at scale of 50% and 150%, respectively.
- In the *rotation* and *shearing* experiments, each image is rotated and sheared at a randomly sampled degree between -30° and 30°.
- For the *Gaussian noise* experiment, noise is sampled once per pixel from a normal distribution $N(0, s)$, where $s$ is sampled between 0 and $0.2 \times 255$; subsequently, Gaussian noise is added to each image.
- For the *Gaussian blur* experiment, we blur each image with a Gaussian kernel with a sigma of 3.0.

# D Extended Experiments

## D.1 Training Time and Memory

The additional training time and memory required when using *Proxy Synthesis* is negligible. Table A indicates that the computation times required for generating synthetics and computing loss are $0.232ms \sim 0.467ms$ and $0.872ms \sim 1.530ms$, respectively, which is trivial. This is because synthetics can be generated and the losses can be computed using simple linear algebra. In terms of memory, Norm-softmax loss creates $N$ features for both embeddings and proxies and a $N \times N$ similarity matrix for the computation of loss. On the other hand, *PS + Norm-softmax* loss creates $(N + \mu \times N)$ features for both embeddings and proxies and a $(N + \mu \times N) \times (N + \mu \times N)$ similarity matrix for the computation of loss. Considering that the proposed method does not need additional forward propagation for generating synthetics unlike previous methods (Zhao et al. 2018; Duan et al. 2018; Zheng et al. 2019), the additional memory required for the generation of synthetics is negligible.

## D.2 Effect of Hyper-parameters

In Figure A, we investigate the effect of the two hyper-parameters $\alpha$ and $\mu$ used in *Proxy Synthesis* by varying the values of $\alpha \in \{0.2, 0.4, 0.8, 1.0, 1.5, 2.0\}$ and $\mu \in \{0.2, 0.5, 1.0, 1.5, 2.0\}$. For $\alpha$ in the interpolation function, smaller values $(0.2 \sim 0.8)$ show relatively good performance because they lead to a higher chance of generating hard synthetic classes near the original classes, as discussed in Section 4.3 from the main paper. Moreover, the results suggest that when the generation ratio $\mu$ is around 1.0, the performance is high. We speculate that this is because too few synthetics by a small value of $\mu$ generate an insufficient effect of *Proxy Synthesis*, whereas too many synthetics by a large value of $\mu$ can be a distraction to learn with information of originals. Thus, a value of $\mu$ close to 1.0, where the number of synthetics is similar to that of the originals, leads to high and stable performance.

| | Train | | Test | | | Train | | Test | |
|---|---|---|---|---|---|---|---|---|---|
| $\lambda$ | NS | PS + NS | NS | PS + NS | $\lambda$ | PN | PS + PN | PN | PS + PN |
| 0.10 | 15.4 | 38.3 (+22.9) | 6.4 | 13.8 (+7.4) | 0.10 | 14.6 | 40.5 (+25.9) | 7.8 | 15.3 (+7.5) |
| 0.15 | 16.9 | 38.4 (+21.5) | 6.8 | 14.1 (+7.4) | 0.15 | 14.8 | 41.6 (+26.8) | 7.9 | 15.9 (+8.0) |
| 0.20 | 16.5 | 39.0 (+22.5) | 7.3 | 14.9 (+7.6) | 0.20 | 14.8 | 40.5 (+25.7) | 8.2 | 16.2 (+8.1) |
| 0.25 | 16.3 | 38.5 (+22.2) | 7.1 | 14.7 (+7.6) | 0.25 | 15.4 | 43.0 (+27.7) | 8.5 | 16.5 (+7.9) |
| 0.30 | 16.7 | 39.7 (+23.0) | 7.3 | 16.5 (+9.2) | 0.30 | 16.2 | 44.2 (+28.0) | 8.6 | 17.5 (+8.9) |
| 0.35 | 18.0 | 42.0 (+24.0) | 7.9 | 17.2 (+9.3) | 0.35 | 16.2 | 45.2 (+29.1) | 9.3 | 19.1 (+9.8) |
| 0.40 | 19.6 | 43.4 (+23.8) | 8.6 | 18.1 (+9.5) | 0.40 | 17.2 | 47.2 (+30.1) | 9.9 | 20.2 (+10.3) |
| 0.45 | 22.7 | 49.0 (+26.4) | 9.6 | 20.2 (+10.6) | 0.45 | 22.2 | 50.4 (+28.1) | 11.5 | 22.4 (+10.9) |
| 0.50 | 37.6 | 62.8 (+25.2) | 13.0 | 24.3 (+11.3) | 0.50 | 37.7 | 60.5 (+22.8) | 16.3 | 24.8 (+8.5) |
| 0.55 | 22.9 | 50.0 (+27.1) | 9.7 | 20.5 (+10.9) | 0.55 | 22.0 | 49.9 (+27.9) | 11.7 | 21.5 (+9.9) |
| 0.60 | 18.3 | 43.6 (+25.3) | 9.1 | 18.7 (+9.6) | 0.60 | 17.4 | 46.2 (+28.8) | 9.9 | 20.2 (+10.2) |
| 0.65 | 17.7 | 41.7 (+24.0) | 7.6 | 17.2 (+9.7) | 0.65 | 15.6 | 45.3 (+29.7) | 9.1 | 17.9 (+8.8) |
| 0.70 | 18.0 | 39.8 (+21.8) | 7.5 | 16.1 (+8.7) | 0.70 | 16.3 | 43.5 (+27.2) | 9.3 | 17.1 (+7.8) |
| 0.75 | 17.3 | 39.2 (+22.0) | 7.4 | 14.7 (+7.3) | 0.75 | 14.6 | 42.6 (+28.0) | 7.6 | 16.7 (+9.1) |
| 0.80 | 16.7 | 40.0 (+23.3) | 6.9 | 14.2 (+7.3) | 0.80 | 14.6 | 41.3 (+26.8) | 8.1 | 15.9 (+7.8) |
| 0.85 | 16.2 | 39.1 (+22.9) | 6.8 | 13.9 (+7.1) | 0.85 | 14.8 | 41.1 (+26.3) | 7.4 | 15.3 (+8.0) |
| 0.90 | 15.7 | 38.3 (+22.6) | 6.5 | 13.7 (+7.2) | 0.90 | 14.8 | 40.5 (+25.6) | 7.1 | 14.9 (+7.8) |
| 0.95 | 15.8 | 38.1 (+22.3) | 6.2 | 14.4 (+8.1) | 0.95 | 14.6 | 40.3 (+25.8) | 7.3 | 15.4 (+8.1) |

(a) Norm-softmax (NS) loss  (b) Proxy-NCA (PN) loss

Table B: Recall@1(%) of embedding deformations with trained networks on CARS196. For a gallery set, synthetic embeddings generated with $\lambda$ and original embeddings are used. For a query set, synthetic embeddings are used to find other synthetic embeddings generated with the same manner.

### D.3 Robustness to Embedding Deformation

In addition to the embedding deformation experiment in Section 4.4, we include additional experimental results on Norm-softmax and Proxy-NCA losses. Table Ba shows performances of Norm-softmax experiment, which is the base values for Figure 5. We conduct the same experiments with Proxy-NCA loss, as shown in Table Bb. For both loss functions, *Proxy Synthesis* shows significant performance gain with a similar performance pattern. It demonstrates that the models can obtain more robust embedding features by *Proxy Synthesis*.

### D.4 Comparison with Other Regularizers

Recent works have studied the different requirements of generalization between classification and metric learning tasks (Roth et al. 2020; Milbich et al. 2020). In the classification task, training and test data share the same class information with i.i.d. training and test distributions; thus, strong focus on the most discriminative directions of significant variance is shown to be advantageous (Verma et al. 2018; Roth et al. 2020). However, for transfer learning, such as metric learning, training data does not share class information with test data. Because of this shift in training and test distribution, recent works (Roth et al. 2020; Milbich et al. 2020) argue that preserving a substantial amount of directions of significant variance is required for generalization in metric learning. The following experiments include the quantitative comparison with Mixup (Zhang et al. 2017; Verma et al. 2018; Guo, Mao, and Zhang 2019) and Virtual softmax (Chen, Deng, and Shen 2018) in classification and retrieval tasks. They demonstrate that both techniques work well for the classification task, while *Proxy Synthesis* works the best for the retrieval task, which verifies the different requirements of generalization between classification and metric learning as studied in (Roth et al. 2020; Milbich et al. 2020).

**Experiment on Classification:** Because softmax loss and its variants are included in proxy-based loss, *Proxy Synthesis* can be used for classification tasks as well. We conduct experiments for comparing the proposed method with Virtual Softmax and Mixup techniques in classification task on four datasets: Cifar10 (Krizhevsky, Hinton et al. 2009), Cifar100 (Krizhevsky, Hinton et al. 2009), SVHN (Netzer et al. 2011), and TinyImagenet (CS231N 2017). We add an implementation of *Proxy Synthesis* in the official implementation [1] of Manifold Mixup, and use the same scripts for every experiment. For a fair comparison, we use the best hyper-parameters reported in (Zhang et al. 2017; Verma et al. 2018) and search for the best hyper-parameters $\alpha$ and $\mu$ of *Proxy Synthesis* as shown in Table C.

The results are presented in Table D. Intriguingly, *Proxy Synthesis* improves the performance of every dataset in the classification task. We speculate that it is because the effect of the smoothing decision boundary gives a positive effect on generalization.

---
[1] https://github.com/vikasverma1077/manifold_mixup

| $\alpha$ \ $\mu$ | 0.25 | 0.50 | 0.75 | 1.00 |
|---|---|---|---|---|
| 0.4 | 95.4 | 95.5 | 95.4 | 95.7 |
| 1.0 | 95.7 | **96.0** | 95.5 | 95.8 |
| 1.5 | 95.9 | 95.7 | 95.6 | 95.8 |
| 2.0 | 95.6 | 95.6 | 95.7 | 95.6 |

(a) Cifar10

| $\alpha$ \ $\mu$ | 0.25 | 0.50 | 0.75 | 1.00 |
|---|---|---|---|---|
| 0.4 | 76.8 | 76.2 | 76.2 | 76.4 |
| 1.0 | 76.6 | 76.8 | 76.4 | 76.5 |
| 1.5 | 76.4 | 76.7 | **77.1** | 76.5 |
| 2.0 | 76.7 | 76.5 | 76.6 | 76.3 |

(b) Cifar100

| $\alpha$ \ $\mu$ | 0.25 | 0.50 | 0.75 | 1.00 |
|---|---|---|---|---|
| 0.4 | 97.4 | 97.2 | 97.3 | **97.6** |
| 1.0 | 97.5 | 97.4 | 97.4 | 97.4 |
| 1.5 | 97.4 | 97.4 | 97.4 | 97.5 |
| 2.0 | 97.5 | 97.5 | 97.5 | 97.5 |

(c) SVHN

| $\alpha$ \ $\mu$ | 0.25 | 0.50 | 0.75 | 1.00 |
|---|---|---|---|---|
| 0.4 | 59.7 | 58.5 | 57.4 | 55.2 |
| 1.0 | 59.8 | 58.2 | 56.8 | 54.4 |
| 1.5 | 59.7 | 59.2 | 56.6 | 52.0 |
| 2.0 | **59.9** | 58.7 | 56.4 | 52.1 |

(d) TinyImagenet

Table C: Accuracy(%) of hyper-parameter search for *Proxy Synthesis* using PreActResNet18 (He et al. 2016b) in image classification task.

| Dataset | Baseline | Virtual Softmax | Input Mixup | Manifold Mixup | *Proxy Synthesis* |
|---|---|---|---|---|---|
| Cifar10 | 95.2 | 95.7 (+0.5) | 96.2 (+1.0) | **97.1 (+1.9)** | 96.0 (+0.8) |
| Cifar100 | 76.0 | 76.8 (+0.8) | 77.9 (+1.9) | **79.7 (+3.7)** | 77.1 (+1.1) |
| SVHN | 97.1 | 97.4 (+0.3) | 97.2 (+0.1) | **97.7 (+0.6)** | 97.6 (+0.5) |
| TinyImagenet | 55.5 | 58.1 (+2.6) | 56.5 (+1.0) | 58.7 (+3.2) | **59.9 (+4.4)** |

Table D: Comparison of accuracy(%) among regularizers in image classification task. With a baseline of softmax loss, we train each model with softmax loss including Virtual Softmax, Input Mixup, Manifold Mixup, and *Proxy Synthesis* using PreActResNet18 (He et al. 2016b).

The gains of *Proxy Synthesis* are always higher than that of Virtual Softmax because Virtual Softmax is limited by generating one synthetic negative class, while *Proxy Synthesis* generates multiple synthetic classes, including synthetic embeddings. However, the performance gain of *Proxy Synthesis* is lower than that of Input and Manifold Mixup except for TinyImagenet. Because *Proxy Synthesis* considers synthetic classes as different classes from the original classes, this can be a distraction for the network training positional confidence of each original class. On the other hand, Input and Manifold Mixup improve performance by considering linear behavior in-between classes, which is more suitable for the classification task.

**Experiment on Retrieval:** To further compare the proposed method with other regularizers, we train networks for image retrieval tasks on Norm-softmax loss with different regularizers. For a fair comparison, we search for the best performing value of $\alpha$ in the Input and Manifold Mixup as shown in Table E, and use $\alpha = 0.4$ and $\mu = 1.0$ for *Proxy Synthesis*. In Virtual Softmax, we observe that $W_{virt}^T X_i$ is much bigger than that of positive class $W_{y_i}^T X_i$ because of the same direction of vectors $W_{virt}$ and $X_i$ and it fails to train the network. To handle these issues, we clip the similarity with a virtual negative class to be under 10 for softmax. Moreover, Virtual Softmax is not applicable to Norm-softmax loss because $W_{virt}$ will always be constant 1. The detailed analysis of comparison with other regularizers in the retrieval tasks is provided in Section 4.5 from the main paper.

### D.5 Visualization of Embedding Space

We present t-SNE visualizations (Maaten and Hinton 2008) of models, which are trained with *PS* + Norm-softmax loss on CUB200 and SOP in Figure B and C, respectively. They show a similar pattern with Figure 3 by placing unseen test embeddings in-between the clusters of train embeddings. With such a pattern, synthetic classes, generated in-between train embeddings, would mimic unseen classes during the training phase. For further analysis, we provide t-SNE visualizations by each epoch of a model, which is trained with *PS* + Norm-softmax loss on CARS196 in Figure D. At the beginning of training in Figure Da, all embeddings are scattered, and the proxies are not differentiable from each other. Because the train embeddings and proxies are positioned randomly, the synthetic embeddings and proxies are generated in the meaningless positions, as depicted in Figure Db. At 30 epochs of training, the train embeddings start forming clusters with further discriminative proxies; thus, synthetic embeddings and proxies start being generated in meaningful positions, as depicted in Figure Dd. These synthetic classes help the network learn with additional training signals by imitating *unseen* classes such as original test embeddings in

|       | CARS196 |          | SOP   |          |
|-------|---------|----------|-------|----------|
| $\alpha$ | Input | Manifold | Input | Manifold |
| 0.3   | 81.0    | 81.3     | 76.9  | 77.3     |
| 0.4   | 80.7    | 81.2     | **77.0** | 77.4  |
| 0.5   | **81.1** | **81.6** | 76.7  | **77.5** |
| 1.0   | 80.8    | 81.1     | 75.9  | 76.4     |
| 1.5   | 80.2    | 80.6     | 75.3  | 76.2     |
| 2.0   | 79.4    | 80.4     | 75.2  | 75.9     |

(a) Softmax loss

|       | CARS196 |          | SOP   |          |
|-------|---------|----------|-------|----------|
| $\alpha$ | Input | Manifold | Input | Manifold |
| 0.3   | 81.8    | 82.7     | 77.4  | 77.5     |
| 0.4   | **82.2** | 83.1    | 77.1  | 77.9     |
| 0.5   | **82.2** | **83.6** | **78.2** | **78.4** |
| 1.0   | 81.9    | 83.1     | 77.0  | 77.6     |
| 1.5   | 81.3    | 82.5     | 76.5  | 77.5     |
| 2.0   | 80.9    | 82.3     | 76.6  | 77.4     |

(b) Norm-softmax loss

Table E: Recall@1 (%) performance of hyper-parameter search for Input and Manifold Mixup with BN-Inception (Ioffe and Szegedy 2015) network in image retrieval task.

Figure Dc. After further training, both the original and synthetic embeddings become more discriminative by forming clusters with distances between classes at the 60th epoch, as depicted in Figure De and Df. Finally, we obtain a well-structured embedding space with well-clustered train and test embeddings at the 150th epoch, as depicted in Figure Dg.

### D.6 Comparison with State-of-the-Art

This section contains the extended comparison with state-of-the-art per dataset in both *conventional* and *MLRC evaluation* protocols. The conventional evaluation includes more Recall@k evaluation and existing methods for Table 6 in the main paper. The MLRC evaluation includes the performance of separated models with a 128-dimensional embedding feature for Table 4 in the main paper. The corresponding tables for each dataset are Table F and G for CUB-200-2011 (Wah et al. 2011), Table H and I for CARS196 (Krause et al. 2013), Table J and K for Standford Online Products (Oh Song et al. 2016), and Table L for In-Shop Clothes (Liu et al. 2016). In conventional evaluation, *Proxy Synthesis* improves the performance of proxy-based losses for most Recall@k except for a minority of cases. Compared to the ensemble, sample generation, and pair-based methods, the proposed method outperforms all cases in every dataset with a large margin. In MLRC evaluation, performances of both concatenated 512-dim and separated 128-dim are increased in every metric and dataset. Considering MLRC evaluation is designed with respect to fairness, the result demonstrates that the proposed method enhances generalization ability for pair-based loss in general.

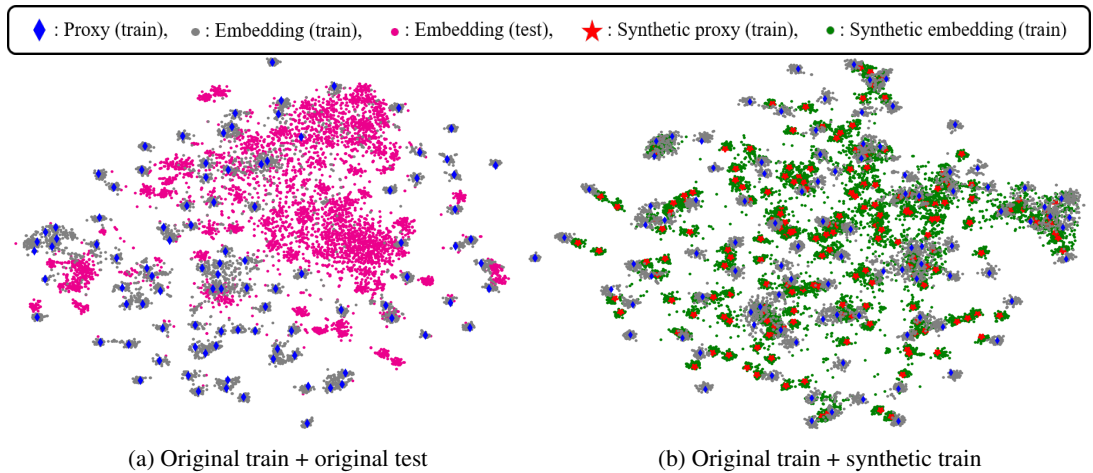

(a) Original train + original test    (b) Original train + synthetic train

Figure B: t-SNE visualization of converged network trained with *PS* + Norm-softmax loss on CUB200. (a) We project both train (*seen*) and test (*unseen*) embeddings. (b) With the same train embeddings as in (a), we project synthetic embeddings and proxies.

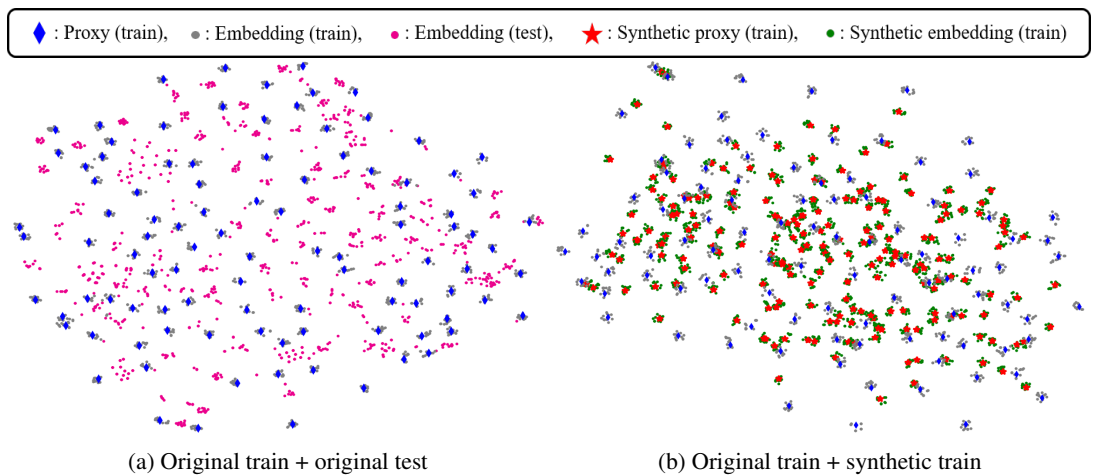

(a) Original train + original test    (b) Original train + synthetic train

Figure C: t-SNE visualization of converged network trained with *PS* + Norm-softmax loss on SOP. As the number of classes is large and the number of instance for each class is small in SOP, we sample 100 classes with more than 10 instances per class for better visualization. (a) We project both train (*seen*) and test (*unseen*) embeddings. (b) With the same train embeddings as in (a), we project synthetic embeddings and proxies.

◆ : Proxy (train),   • : Embedding (train),   • : Embedding (test),   ★ : Synthetic proxy (train),   • : Synthetic embedding (train)

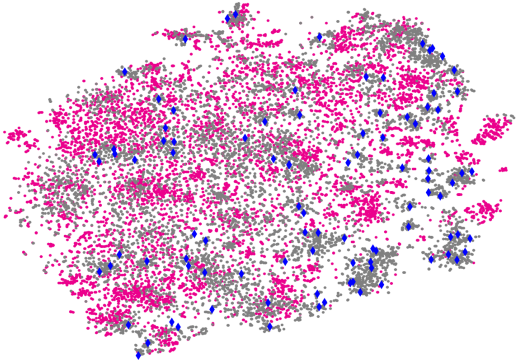
(a) Original train + original test, 2nd epoch

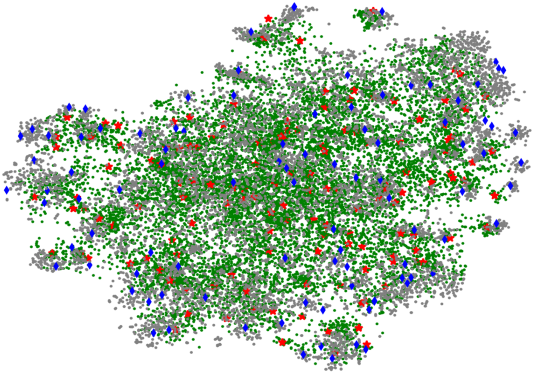
(b) Original train + synthetic train, 2nd epoch

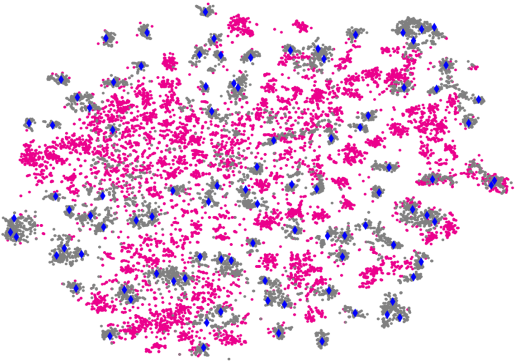
(c) Original train + original test, 30th epoch

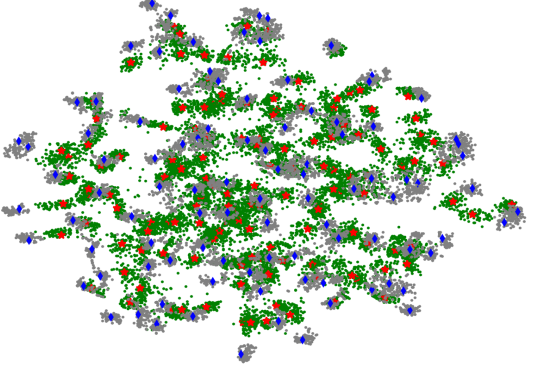
(d) Original train + synthetic train, 30th epoch

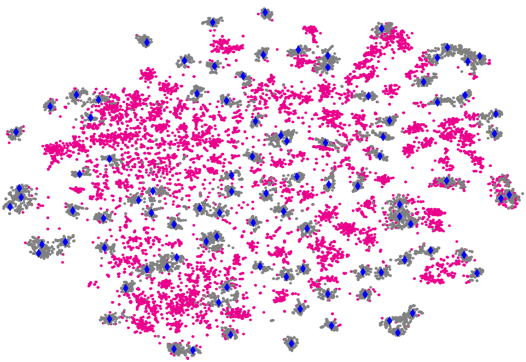
(e) Original train + original test, 60th epoch

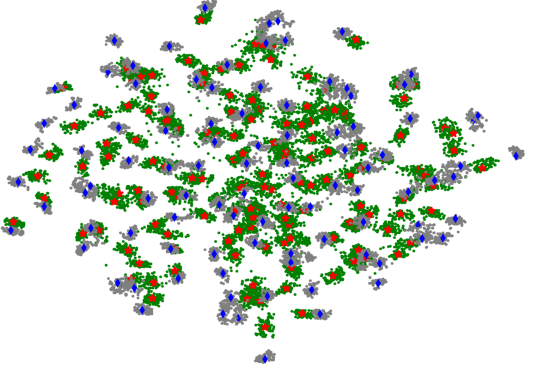
(f) Original train + synthetic train, 60th epoch

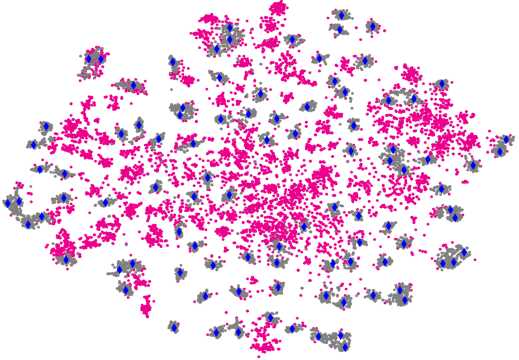
(g) Original train + original test, 150th epoch

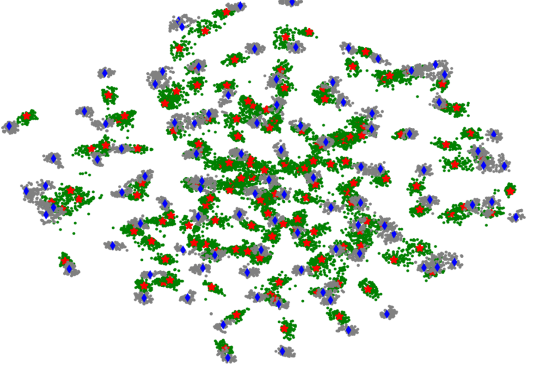
(h) Original train + synthetic train, 150th epoch

Figure D: t-SNE visualization of *PS* + Norm-softmax loss on CARS196 at each epoch. (a, c, e, g) Original train and test data. (b, d, f, h) Original and synthetic train data.

| T | Method | Net | Dim | R@1 | | R@2 | | R@4 | | R@8 | |
|---|---|---|---|---|---|---|---|---|---|---|---|
| | | | | CUB-200-2011 (Wah et al. 2011) | | | | | | | |
| Ens | HDC (Yuan, Yang, and Zhang 2017) | G | 384 | 53.6 | - | 65.7 | - | 77.0 | - | 85.6 | - |
| | A-BIER (Opitz et al. 2018) | G | 512 | 57.5 | - | 68.7 | - | 78.3 | - | 86.2 | - |
| | ABE (Kim et al. 2018) | G | 512 | 60.6 | - | 71.5 | - | 79.8 | - | 87.4 | - |
| Gen | DAML (N-pair) (Duan et al. 2018) | G | 512 | 52.7 | - | 65.4 | - | 75.5 | - | 84.3 | - |
| | HDML (N-pair) (Zheng et al. 2019) | G | 512 | 53.7 | - | 65.7 | - | 76.7 | - | 85.7 | - |
| | Symm+N-pair (Gu and Ko 2020) | G | 512 | 55.9 | - | 67.6 | - | 78.3 | - | 86.2 | - |
| | EE+Multi-Similarity (Ko and Gu 2020) | G | 512 | 57.4 | - | 68.7 | - | 79.5 | - | 86.9 | - |
| Pair | Margin (Wu et al. 2017) | R50 | 128 | 63.6 | - | 74.4 | - | 83.1 | - | 90.0 | - |
| | HTL (Ge 2018) | BN | 512 | 57.1 | - | 68.8 | - | 78.7 | - | 86.5 | - |
| | RLL-H (Wang et al. 2019b) | BN | 512 | 57.4 | - | 69.7 | - | 79.2 | - | 86.9 | - |
| | Multi-Similarity (Wang et al. 2019a) | BN | 512 | 65.7 | - | 77.0 | - | 86.3 | - | 91.2 | - |
| Proxy | Softmax | BN | 512 | 64.2 | - | 75.7 | - | 84.1 | - | 89.9 | - |
| | *PS* + Softmax | BN | 512 | **64.9** | (+0.7) | **76.0** | (+0.3) | **84.3** | (+0.2) | **90.6** | (+0.7) |
| | Norm-softmax (Wang et al. 2017) | BN | 512 | 64.9 | - | 75.7 | - | 84.3 | - | 90.5 | - |
| | *PS* + Norm-softmax | BN | 512 | **66.0** | (+1.1) | **76.6** | (+0.9) | **85.0** | (+0.7) | **90.8** | (+0.3) |
| | SphereFace (Liu et al. 2017) | BN | 512 | 65.4 | - | 76.5 | - | 84.6 | - | **90.8** | - |
| | *PS* + SphereFace | BN | 512 | **66.6** | (+1.2) | **76.6** | (+0.1) | **84.7** | (+0.1) | 90.4 | (-0.4) |
| | Cosface (Wang et al. 2018) | BN | 512 | 65.7 | - | 76.2 | - | **84.7** | - | 90.6 | - |
| | *PS* + Cosface | BN | 512 | **66.6** | (+0.9) | **76.8** | (+0.6) | 84.6 | (-0.1) | **90.7** | (+0.1) |
| | Arcface (Deng et al. 2019) | BN | 512 | 66.1 | - | 76.6 | - | 84.8 | - | 90.7 | - |
| | *PS* + Arcface | BN | 512 | **66.8** | (+0.7) | **77.4** | (+0.8) | **85.0** | (+0.2) | 90.7 | (0.0) |
| | Proxy-NCA (Movshovitz-Attias et al. 2017) | BN | 512 | 65.1 | - | 76.1 | - | **85.0** | - | **90.7** | - |
| | *PS* + Proxy-NCA | BN | 512 | **66.4** | (+1.3) | **76.8** | (+0.7) | 85.0 | (0.0) | 90.6 | (-0.1) |
| | SoftTriple (Qian et al. 2019) | BN | 512 | 65.4 | - | 76.4 | - | 84.5 | - | 90.4 | - |
| | *PS* + SoftTriple | BN | 512 | **66.6** | (+1.2) | **76.8** | (+0.4) | **85.1** | (+0.6) | **90.6** | (+0.2) |
| | Proxy-anchor (Kim et al. 2020) | BN | 512 | 68.4 | - | 79.2 | - | 86.8 | - | 91.6 | - |
| | *PS* + Proxy-anchor | BN | 512 | **69.2** | (+0.8) | **79.5** | (+0.3) | **87.2** | (+0.4) | **91.8** | (+0.2) |
| - | *PS*: Average boost | - | - | - | (+1.0) | - | (+0.5) | - | (+0.3) | - | (+0.1) |
| | *PS*: Minimum boost | - | - | - | (+0.7) | - | (+0.1) | - | (-0.1) | - | (-0.4) |
| | *PS*: Maximum boost | - | - | - | (+1.3) | - | (+0.9) | - | (+0.7) | - | (+0.7) |

Table F: **[Conventional evaluation]** Recall@k (%) on CUB-200-2011 dataset in image retrieval. Method type (T) is denoted by abbreviations: Ens for ensemble methods, Gen for sample generation methods for pair-based losses, Pair for pair-based losses, and Proxy for proxy-based losses. Backbone network (Net) also is denoted by abbreviations: G for GoogleNet (Szegedy et al. 2015), R50 for ResNet50 (He et al. 2016a) and BN for Inception with batch normalization (Ioffe and Szegedy 2015).

| CUB200 | Concatenated (512-dim) | | | Separated (128-dim) | | |
|---|---|---|---|---|---|---|
| Loss | P@1 | RP | MAP@R | P@1 | RP | MAP@R |
| Norm-softmax | 65.65 ± 0.30 | 35.99 ± 0.15 | 25.25 ± 0.13 | 58.75 ± 0.19 | 31.75 ± 0.12 | 20.96 ± 0.11 |
| *PS* + Norm-softmax | **69.19 ± 0.34** | **37.32 ± 0.29** | **26.40 ± 0.29** | **60.17 ± 0.43** | **32.05 ± 0.28** | **21.15 ± 0.26** |
| CosFace | 67.32 ± 0.32 | 37.49 ± 0.21 | 26.70 ± 0.23 | 59.63 ± 0.36 | 31.99 ± 0.22 | 21.21 ± 0.22 |
| *PS* + CosFace | **69.52 ± 0.26** | **37.99 ± 0.23** | **27.10 ± 0.23** | **60.97 ± 0.32** | **32.55 ± 0.19** | **21.62 ± 0.17** |
| ArcFace | 67.50 ± 0.25 | 37.31 ± 0.21 | 26.45 ± 0.20 | 60.17 ± 0.32 | 32.37 ± 0.17 | 21.49 ± 0.16 |
| *PS* + ArcFace | **68.79 ± 0.31** | **37.46 ± 0.26** | **26.79 ± 0.27** | **60.83 ± 0.38** | **32.52 ± 0.28** | **21.70 ± 0.27** |
| SoftTriple | 67.73 ± 0.39 | 37.34 ± 0.19 | 26.51 ± 0.20 | 59.94 ± 0.33 | 32.12 ± 0.14 | 21.31 ± 0.14 |
| *PS* + SoftTriple | **68.26 ± 0.16** | **37.98 ± 0.21** | **27.02 ± 0.21** | **60.35 ± 0.26** | **32.40 ± 0.11** | **21.51 ± 0.12** |
| Proxy-NCA | 65.69 ± 0.43 | 35.14 ± 0.26 | 24.21 ± 0.27 | 57.88 ± 0.30 | 30.16 ± 0.22 | 19.32 ± 0.21 |
| *PS* + Proxy-NCA | **66.02 ± 0.29** | **35.73 ± 0.24** | **24.84 ± 0.22** | **58.52 ± 0.24** | **30.86 ± 0.15** | **20.08 ± 0.15** |
| Proxy-anchor | 69.73 ± 0.31 | 38.23 ± 0.37 | 27.44 ± 0.35 | 61.50 ± 0.34 | 32.94 ± 0.25 | 22.19 ± 0.25 |
| *PS* + Proxy-anchor | **70.41 ± 0.36** | **38.82 ± 0.29** | **28.11 ± 0.29** | **61.91 ± 0.21** | **33.27 ± 0.21** | **22.41 ± 0.19** |

Table G: **[MLRC evaluation]** Performance (%) on CUB-200-2011 dataset in image retrieval. We report the performance of concatenated 512-dim and separated 128-dim. Bold numbers indicate the best score within the same loss.

| T | Method | Net | Dim | R@1 | | R@2 | | R@4 | | R@8 | |
|---|---|---|---|---|---|---|---|---|---|---|---|
| | | | CARS196 (Krause et al. 2013) | | | | | | | | |
| Ens | HDC (Yuan, Yang, and Zhang 2017) | G | 384 | 73.7 | - | 83.2 | - | 89.5 | - | 93.8 | - |
| | A-BIER (Opitz et al. 2018) | G | 512 | 82.0 | - | 89.0 | - | 93.2 | - | 96.1 | - |
| | ABE (Kim et al. 2018) | G | 512 | 85.2 | - | 90.5 | - | 94.0 | - | 96.1 | - |
| Gen | DAML (N-pair) (Duan et al. 2018) | G | 512 | 75.1 | - | 83.8 | - | 89.7 | - | 93.5 | - |
| | HDML (N-pair) (Zheng et al. 2019) | G | 512 | 79.1 | - | 87.1 | - | 92.1 | - | 95.5 | - |
| | Symm+N-pair (Gu and Ko 2020) | G | 512 | 76.5 | - | 84.3 | - | 90.4 | - | 94.1 | - |
| | EE+Multi-Similarity (Ko and Gu 2020) | G | 512 | 76.1 | - | 84.2 | - | 89.8 | - | 93.8 | - |
| Pair | Margin (Wu et al. 2017) | R50 | 128 | 79.6 | - | 86.5 | - | 91.9 | - | 95.1 | - |
| | HTL (Ge 2018) | BN | 512 | 81.4 | - | 88.0 | - | 92.7 | - | 95.7 | - |
| | RLL-H (Wang et al. 2019b) | BN | 512 | 74.0 | - | 83.6 | - | 90.1 | - | 94.1 | - |
| | Multi-Similarity (Wang et al. 2019a) | BN | 512 | 84.1 | - | 90.4 | - | 94.0 | - | 96.5 | - |
| Proxy | Softmax | BN | 512 | 81.5 | - | 89.0 | - | 93.6 | - | 96.8 | - |
| | *PS* + Softmax | BN | 512 | **84.3** | (+2.8) | **90.5** | (+1.5) | **94.6** | (+1.0) | **97.0** | (+0.2) |
| | Norm-softmax (Wang et al. 2017) | BN | 512 | 83.3 | - | 89.7 | - | 94.1 | - | 96.7 | - |
| | *PS* + Norm-softmax | BN | 512 | **84.7** | (+1.4) | **90.7** | (+1.0) | **94.6** | (+0.5) | **96.9** | (+0.2) |
| | SphereFace (Liu et al. 2017) | BN | 512 | 83.6 | - | 90.5 | - | 94.3 | - | 96.9 | - |
| | *PS* + SphereFace | BN | 512 | **85.1** | (+1.5) | **91.0** | (+0.5) | **94.6** | (+0.3) | **97.1** | (+0.2) |
| | Cosface (Wang et al. 2018) | BN | 512 | 83.6 | - | 89.9 | - | 94.2 | - | 96.6 | - |
| | *PS* + Cosface | BN | 512 | **84.6** | (+1.0) | **90.8** | (+0.9) | **94.3** | (+0.1) | **96.8** | (+0.2) |
| | Arcface (Deng et al. 2019) | BN | 512 | 83.7 | - | 90.0 | - | 94.3 | - | 96.8 | - |
| | *PS* + Arcface | BN | 512 | **84.7** | (+1.0) | **91.0** | (+1.0) | **94.8** | (+0.5) | **97.0** | (+0.2) |
| | Proxy-NCA (Movshovitz-Attias et al. 2017) | BN | 512 | 83.7 | - | 90.4 | - | 94.1 | - | 96.9 | - |
| | *PS* + Proxy-NCA | BN | 512 | **84.5** | (+0.8) | **90.8** | (+0.4) | **94.4** | (+0.3) | **97.0** | (+0.1) |
| | SoftTriple (Qian et al. 2019) | BN | 512 | 84.5 | - | 90.7 | - | 94.5 | - | 96.9 | - |
| | *PS* + SoftTriple | BN | 512 | **85.3** | (+0.8) | **91.0** | (+0.3) | **94.8** | (+0.3) | **97.1** | (+0.2) |
| | Proxy-anchor (Kim et al. 2020) | BN | 512 | 86.1 | - | 91.7 | - | 95.0 | - | 97.3 | - |
| | *PS* + Proxy-anchor | BN | 512 | **86.9** | (+0.8) | **92.4** | (+0.7) | **95.2** | (+0.2) | 97.3 | (0.0) |
| - | *PS*: Average boost | - | - | - | (+1.3) | - | (+0.8) | - | (+0.4) | - | (+0.2) |
| | *PS*: Minimum boost | - | - | - | (+0.8) | - | (+0.3) | - | (+0.1) | - | (0.0) |
| | *PS*: Maximum boost | - | - | - | (+2.8) | - | (+1.5) | - | (+1.0) | - | (+0.2) |

Table H: **[Conventional evaluation]** Recall@k (%) on CARS196 dataset in image retrieval. Method type (T) is denoted by abbreviations: Ens for ensemble methods, Gen for sample generation methods for pair-based losses, Pair for pair-based losses, and Proxy for proxy-based losses. Backbone network (Net) also is denoted by abbreviations: G for GoogleNet (Szegedy et al. 2015), R50 for ResNet50 (He et al. 2016a) and BN for Inception with batch normalization (Ioffe and Szegedy 2015).

| CARS196 | Concatenated (512-dim) | | | Separated (128-dim) | | |
|---|---|---|---|---|---|---|
| Loss | P@1 | RP | MAP@R | P@1 | RP | MAP@R |
| Norm-softmax | 83.16 ± 0.25 | 36.20 ± 0.26 | 26.00 ± 0.30 | 72.55 ± 0.18 | 29.35 ± 0.20 | 18.73 ± 0.20 |
| *PS* + Norm-softmax | **85.70 ± 0.24** | **38.33 ± 0.31** | **28.31 ± 0.32** | **75.86 ± 0.30** | **31.28 ± 0.36** | **20.61 ± 0.37** |
| CosFace | 85.52 ± 0.24 | 37.32 ± 0.28 | 27.57 ± 0.30 | 74.67 ± 0.20 | 29.01 ± 0.11 | 18.80 ± 0.12 |
| *PS* + CosFace | **85.58 ± 0.27** | **38.01 ± 0.19** | **27.89 ± 0.20** | **76.21 ± 0.33** | **31.40 ± 0.33** | **20.76 ± 0.35** |
| ArcFace | 85.44 ± 0.28 | 37.02 ± 0.29 | 27.22 ± 0.30 | 72.10 ± 0.37 | 27.29 ± 0.17 | 17.11 ± 0.18 |
| *PS* + ArcFace | **85.59 ± 0.25** | **38.31 ± 0.22** | **28.24 ± 0.20** | **75.22 ± 0.42** | **30.79 ± 0.38** | **20.18 ± 0.38** |
| SoftTriple | 84.49 ± 0.26 | 37.03 ± 0.21 | 28.07 ± 0.21 | 73.69 ± 0.21 | 29.29 ± 0.16 | 19.32 ± 0.18 |
| *PS* + SoftTriple | **85.53 ± 0.12** | **38.40 ± 0.20** | **28.45 ± 0.19** | **75.82 ± 0.40** | **31.59 ± 0.27** | **20.99 ± 0.25** |
| Proxy-NCA | 83.56 ± 0.27 | 35.62 ± 0.28 | 25.38 ± 0.31 | 73.46 ± 0.23 | 28.90 ± 0.22 | 18.29 ± 0.22 |
| *PS* + Proxy-NCA | **84.61 ± 0.19** | **36.39 ± 0.25** | **26.04 ± 0.27** | **75.04 ± 0.39** | **29.88 ± 0.32** | **19.20 ± 0.32** |
| Proxy-anchor | 86.20 ± 0.21 | 39.08 ± 0.31 | 29.37 ± 0.29 | 76.97 ± 0.40 | 31.71 ± 0.53 | 21.29 ± 0.56 |
| *PS* + Proxy-anchor | **86.90 ± 0.35** | **39.38 ± 0.27** | **29.71 ± 0.25** | **77.16 ± 0.39** | **31.90 ± 0.43** | **21.47 ± 0.46** |

Table I: **[MLRC evaluation]** Performance (%) on CARS196 dataset in image retrieval. We report the performance of concatenated 512-dim and separated 128-dim. Bold numbers indicate the best score within the same loss.

| T | Method | Net | Dim | R@1 | | R@10 | | R@100 | | R@1000 | |
|---|---|---|---|---|---|---|---|---|---|---|---|
| | | | | Standford Online Products (Oh Song et al. 2016) | | | | | | | |
| Ens | HDC (Yuan, Yang, and Zhang 2017) | G | 384 | 69.5 | - | 84.4 | - | 92.8 | - | 97.7 | - |
| | A-BIER (Opitz et al. 2018) | G | 512 | 74.2 | - | 86.9 | - | 94.0 | - | 97.8 | - |
| | ABE (Kim et al. 2018) | G | 512 | 76.3 | - | 88.4 | - | 94.8 | - | 98.2 | - |
| Gen | DAML (N-pair) (Duan et al. 2018) | G | 512 | 68.4 | - | 83.5 | - | 92.3 | - | - | - |
| | HDML (N-pair) (Zheng et al. 2019) | G | 512 | 68.7 | - | 83.2 | - | 92.4 | - | - | - |
| | Symm+N-pair (Gu and Ko 2020) | G | 512 | 73.2 | - | 86.7 | - | 94.8 | - | - | - |
| | EE+Multi-Similarity (Ko and Gu 2020) | G | 512 | 78.1 | - | 90.3 | - | 95.8 | - | - | - |
| Pair | Margin (Wu et al. 2017) | R50 | 128 | 72.7 | - | 86.2 | - | 93.8 | - | 98.0 | - |
| | HTL (Ge 2018) | BN | 512 | 74.8 | - | 88.3 | - | 94.8 | - | 98.4 | - |
| | RLL-H (Wang et al. 2019b) | BN | 512 | 76.1 | - | 89.1 | - | 95.4 | - | - | - |
| | Multi-Similarity (Wang et al. 2019a) | BN | 512 | 78.2 | - | 90.5 | - | 96.0 | - | 98.7 | - |
| Proxy | Softmax | BN | 512 | 76.3 | - | 88.5 | - | 94.8 | - | 98.1 | - |
| | *PS* + Softmax | BN | 512 | **77.6** | (+1.3) | **89.3** | (+0.8) | **95.3** | (+0.5) | **98.4** | (+0.3) |
| | Norm-softmax (Wang et al. 2017) | BN | 512 | 78.6 | - | 90.5 | - | 96.0 | - | 98.6 | - |
| | *PS* + Norm-softmax | BN | 512 | **79.6** | (+1.0) | **90.9** | (+0.4) | **96.2** | (+0.2) | **98.7** | (+0.1) |
| | SphereFace (Liu et al. 2017) | BN | 512 | 78.9 | - | 90.6 | - | 95.8 | - | 98.5 | - |
| | *PS* + SphereFace | BN | 512 | **79.4** | (+0.5) | **90.7** | (+0.1) | **96.2** | (+0.4) | **98.8** | (+0.3) |
| | Cosface (Wang et al. 2018) | BN | 512 | 78.6 | - | 90.4 | - | 95.8 | - | 98.5 | - |
| | *PS* + Cosface | BN | 512 | **79.3** | (+0.7) | **90.7** | (+0.3) | **95.9** | (+0.1) | **98.6** | (+0.1) |
| | Arcface (Deng et al. 2019) | BN | 512 | 78.8 | - | 90.5 | - | 95.9 | - | 98.6 | - |
| | *PS* + Arcface | BN | 512 | **79.7** | (+0.9) | **90.9** | (+0.4) | **96.1** | (+0.2) | **98.7** | (+0.1) |
| | Proxy-NCA (Movshovitz-Attias et al. 2017) | BN | 512 | 78.1 | - | 90.0 | - | 95.9 | - | **98.7** | - |
| | *PS* + Proxy-NCA | BN | 512 | **79.1** | (+1.0) | **90.6** | (+0.6) | 95.9 | (0.0) | 98.6 | (-0.1) |
| | SoftTriple (Qian et al. 2019) | BN | 512 | 78.3 | - | 90.4 | - | 96.0 | - | 98.3 | - |
| | *PS* + SoftTriple | BN | 512 | **79.5** | (+1.2) | **90.6** | (+0.2) | 96.0 | (0.0) | **98.6** | (+0.3) |
| | Proxy-anchor (Kim et al. 2020) | BN | 512 | 79.1 | - | **90.8** | - | 96.2 | - | 98.7 | - |
| | *PS* + Proxy-anchor | BN | 512 | **79.8** | (+0.7) | **90.9** | (+0.1) | **96.4** | (+0.2) | **98.8** | (+0.1) |
| - | *PS*: Average boost | - | - | - | (+0.9) | - | (+0.4) | - | (+0.2) | - | (+0.2) |
| | *PS*: Minimum boost | - | - | - | (+0.5) | - | (+0.1) | - | (0.0) | - | (-0.1) |
| | *PS*: Maximum boost | - | - | - | (+1.3) | - | (+0.8) | - | (+0.5) | - | (+0.3) |

Table J: [**Conventional evaluation**] Recall@k (%) on SOP dataset in image retrieval. Method type (T) is denoted by abbreviations: Ens for ensemble methods, Gen for sample generation methods for pair-based losses, Pair for pair-based losses, and Proxy for proxy-based losses. Backbone network (Net) also is denoted by abbreviations: G for GoogleNet (Szegedy et al. 2015), R50 for ResNet50 (He et al. 2016a) and BN for Inception with batch normalization (Ioffe and Szegedy 2015).

| SOP | Concatenated (512-dim) | | | Separated (128-dim) | | |
|---|---|---|---|---|---|---|
| Loss | P@1 | RP | MAP@R | P@1 | RP | MAP@R |
| Norm-softmax | 75.67 ± 0.17 | 50.01 ± 0.22 | 47.13 ± 0.22 | 71.65 ± 0.14 | 45.32 ± 0.17 | 42.35 ± 0.16 |
| *PS* + Norm-softmax | **76.73 ± 0.15** | **51.46 ± 0.21** | **48.52 ± 0.20** | **72.92 ± 0.13** | **46.86 ± 0.13** | **43.84 ± 0.13** |
| CosFace | 75.79 ± 0.14 | 49.77 ± 0.19 | 46.92 ± 0.19 | 70.71 ± 0.19 | 43.56 ± 0.21 | 40.69 ± 0.21 |
| *PS* + CosFace | **76.89 ± 0.20** | **51.60 ± 0.31** | **48.68 ± 0.33** | **73.03 ± 0.15** | **46.90 ± 0.16** | **43.90 ± 0.15** |
| ArcFace | 76.20 ± 0.27 | 50.27 ± 0.38 | 47.41 ± 0.40 | 70.88 ± 1.51 | 44.00 ± 1.26 | 41.11 ± 1.22 |
| *PS* + ArcFace | **77.21 ± 0.20** | **51.90 ± 0.23** | **49.02 ± 0.21** | **73.08 ± 0.15** | **46.74 ± 0.17** | **43.74 ± 0.17** |
| SoftTriple | 76.12 ± 0.17 | 50.21 ± 0.18 | 47.35 ± 0.19 | 70.88 ± 0.20 | 43.83 ± 0.20 | 40.92 ± 0.20 |
| *PS* + SoftTriple | **77.59 ± 0.26** | **52.45 ± 0.21** | **49.53 ± 0.23** | **73.64 ± 0.17** | **47.58 ± 0.19** | **44.55 ± 0.19** |
| Proxy-NCA | 75.89 ± 0.17 | 50.10 ± 0.22 | 47.22 ± 0.21 | 71.30 ± 0.20 | 44.71 ± 0.21 | 41.74 ± 0.21 |
| *PS* + Proxy-NCA | **76.78 ± 0.21** | **51.39 ± 0.27** | **48.44 ± 0.27** | **72.81 ± 0.16** | **46.63 ± 0.19** | **46.59 ± 0.18** |
| Proxy-anchor | 75.37 ± 0.15 | 50.19 ± 0.14 | 47.25 ± 0.15 | 71.56 ± 0.11 | 46.13 ± 0.21 | 43.03 ± 0.21 |
| *PS* + Proxy-anchor | **75.52 ± 0.21** | **50.45 ± 0.22** | **47.49 ± 0.20** | **71.97 ± 0.16** | **46.41 ± 0.17** | **43.35 ± 0.19** |

Table K: [**MLRC evaluation**] Performance (%) on Standford Online Products dataset in image retrieval. We report the performance of concatenated 512-dim and separated 128-dim. Bold numbers indicate the best score within the same loss.

| T | Method | Net | Dim | R@1 | | R@10 | | R@20 | | R@40 | |
|---|---|---|---|---|---|---|---|---|---|---|---|
| | | | | In-Shop Clothes (Liu et al. 2016) | | | | | | | |
| Ens | HDC (Yuan, Yang, and Zhang 2017) | G | 384 | 62.1 | - | 84.9 | - | 89.0 | - | 92.3 | - |
| | A-BIER (Opitz et al. 2018) | G | 512 | 83.1 | - | 95.1 | - | 96.9 | - | 97.8 | - |
| | ABE (Kim et al. 2018) | G | 512 | 87.3 | - | 96.7 | - | 97.9 | - | 98.5 | - |
| Pair | FashionNet (Liu et al. 2016) | V16 | 4096 | 53.0 | - | 73.0 | - | 76.0 | - | 79.0 | - |
| | HTL (Ge 2018) | BN | 128 | 80.9 | - | 94.3 | - | 95.8 | - | 97.4 | - |
| | Multi-Similarity (Wang et al. 2019a) | BN | 512 | 89.7 | - | 97.9 | - | 98.5 | - | 99.1 | - |
| Proxy | Softmax | BN | 512 | 90.4 | - | 97.8 | - | 98.4 | - | 99.0 | - |
| | *PS* + Softmax | BN | 512 | **90.9** | (+0.5) | **97.9** | (+0.1) | **98.5** | (+0.1) | **99.2** | (+0.2) |
| | Norm-softmax (Wang et al. 2017) | BN | 512 | 90.4 | - | 97.7 | - | 98.5 | - | 98.9 | - |
| | *PS* + Norm-softmax | BN | 512 | **91.5** | (+1.1) | **98.1** | (+0.4) | **98.7** | (+0.2) | **99.1** | (+0.2) |
| | SphereFace (Liu et al. 2017) | BN | 512 | 90.3 | - | 97.6 | - | 98.4 | - | 98.9 | - |
| | *PS* + SphereFace | BN | 512 | **91.6** | (+1.3) | **98.0** | (+0.4) | **98.7** | (+0.3) | **99.1** | (+0.2) |
| | Cosface (Wang et al. 2018) | BN | 512 | 90.7 | - | 97.6 | - | 98.3 | - | 98.8 | - |
| | *PS* + Cosface | BN | 512 | **91.4** | (+0.7) | **97.8** | (+0.2) | **98.5** | (+0.2) | **99.0** | (+0.2) |
| | Arcface (Deng et al. 2019) | BN | 512 | 91.0 | - | 97.7 | - | 98.4 | - | 98.9 | - |
| | *PS* + Arcface | BN | 512 | **91.7** | (+0.7) | **98.1** | (+0.4) | **98.7** | (+0.3) | **99.1** | (+0.2) |
| | Proxy-NCA (Movshovitz-Attias et al. 2017) | BN | 512 | 90.0 | - | 97.7 | - | 98.4 | - | 99.0 | - |
| | *PS* + Proxy-NCA | BN | 512 | **91.4** | (+1.4) | **98.0** | (+0.3) | **98.7** | (+0.3) | **99.2** | (+0.2) |
| | SoftTriple (Qian et al. 2019) | BN | 512 | 91.1 | - | 97.8 | - | 98.4 | - | 98.9 | - |
| | *PS* + SoftTriple | BN | 512 | **91.8** | (+0.7) | **98.1** | (+0.3) | **98.7** | (+0.3) | **99.1** | (+0.2) |
| | Proxy-anchor (Kim et al. 2020) | BN | 512 | 91.5 | - | 98.1 | - | 98.8 | - | 99.1 | - |
| | *PS* + Proxy-anchor | BN | 512 | **91.9** | (+0.4) | **98.2** | (+0.1) | 98.8 | (0.0) | 99.1 | (0.0) |
| - | *PS*: Average boost | - | - | - | (+0.9) | - | (+0.3) | - | (+0.2) | - | (+0.2) |
| | *PS*: Minimum boost | - | - | - | (+0.4) | - | (+0.1) | - | (0.0) | - | (0.0) |
| | *PS*: Maximum boost | - | - | - | (+1.4) | - | (+0.4) | - | (+0.3) | - | (+0.2) |

Table L: **[Conventional evaluation]** Recall@k (%) on In-Shop Clothes dataset in image retrieval. Method type (T) is denoted by abbreviations: Ens for ensemble methods, Pair for pair-based losses, and Proxy for proxy-based losses. Backbone network (Net) also is denoted by abbreviations: G for GoogleNet (Szegedy et al. 2015), V16 for VGG16 (Simonyan and Zisserman 2014) and BN for Inception with batch normalization (Ioffe and Szegedy 2015).



\end{document}